\documentclass{article}
\usepackage{ijcai25}

\usepackage[svgnames]{xcolor}
\colorlet{N}{blue!70}
\colorlet{O}{red!70}
\colorlet{Cl}{green!70}
\colorlet{S}{yellow!70}
\colorlet{F}{green!30!yellow!70}
\colorlet{Fe}{orange!50!black!70}
\colorlet{r}{magenta!70}
\colorlet{rt}{r!30}

\usepackage{times}
\usepackage{soul}
\usepackage{url}
\usepackage[hidelinks]{hyperref}
\usepackage[utf8]{inputenc}
\usepackage[small]{caption}
\usepackage{graphicx}
\usepackage{amsmath}
\usepackage{amsthm}
\usepackage{amsfonts}
\usepackage{booktabs}
\usepackage{algorithm}
\usepackage{algpseudocode}
\usepackage[switch]{lineno}
\usepackage{subcaption}
\usepackage{natbib}
\usepackage{mathtools}
\usepackage{float}
\usepackage{siunitx}
\usepackage[capitalise,noabbrev]{cleveref}
\usepackage{tikz}
\usetikzlibrary{shapes, positioning, calc, decorations}
\usepackage{chemfig}
\usepackage{listings}
\setchemfig{
    angle increment=30,
    atom sep=1.67em,
    double bond sep=0.67ex,
    bond style={line width=0.1em},
    cram width=0.8ex,
    cram dash width=0.1em,
    cram dash sep=0.2em,
    arrow style={line width=0.067em},
    arrow head=-{Triangle},
    arrow label sep=1ex,
    cycle radius coeff=0.75,
    chemfig style={line width=0.1em},
}
\setcharge{
    extra sep = 0.5ex,
    .radius = 0.05em,
    .style = {fill = fg},
}

\newcommand\mdoubleplus{\mathbin{+\mkern-5mu+}}

\title{Diversity-Aware Reinforcement Learning for \textit{de novo} Drug Design}

\author{
Hampus Gummesson Svensson$^{1,2}$\and
Christian Tyrchan$^3$\and
Ola Engkvist$^{1,2}$\And
Morteza Haghir Chehreghani$^1$\\
\affiliations
$^1$Department of Computer Science and Engineering, Chalmers University of Technology and University of Gothenburg, Gothenburg, Sweden\\
$^2$Molecular AI, Discovery Sciences, R\&D, AstraZeneca, Gothenburg, Sweden\\
$^3$Medicinal Chemistry, Research and Early Development, Respiratory and Immunology (R\&I), BioPharmaceuticals R\&D, AstraZeneca, Gothenburg, Sweden\\
\emails
hamsven@chalmers.se
}

\begin{document}

\maketitle 

\begin{abstract}
Fine-tuning a pre-trained generative model has demonstrated good performance in generating promising drug molecules. The fine-tuning task is often formulated as a reinforcement learning problem, where previous methods efficiently learn to optimize a reward function to generate potential drug molecules. Nevertheless, in the absence of an adaptive update mechanism for the reward function, the optimization process can become stuck in local optima. The efficacy of the optimal molecule in a local optimization may not translate to usefulness in the subsequent drug optimization process or as a potential standalone clinical candidate. Therefore, it is important to generate a diverse set of promising molecules. Prior work has modified the reward function by penalizing structurally similar molecules, primarily focusing on finding molecules with higher rewards. 
To date, no study has comprehensively examined how different adaptive update mechanisms for the reward function influence the diversity of generated molecules. In this work, we investigate a wide range of intrinsic motivation methods and strategies to penalize the extrinsic reward, and how they affect the diversity of the set of generated molecules. Our experiments reveal that combining structure- and prediction-based methods generally yields better results in terms of diversity.
\end{abstract}


\section{Introduction}
\begin{figure*}
    \centering
    \resizebox{0.89\hsize}{!}{%
    \newsavebox{\MoleculeBox}
\begin{lrbox}{\MoleculeBox}
\chemname[2ex]{\chemfig{
    \textcolor{O}{O}=[1]
    *6(
        -\textcolor{N}{N}(-)
        -(*5(
            -\textcolor{N}{N}
            =(-\textcolor{Cl}{Cl})
            -\chemabove{\textcolor{N}{N}}{\textcolor{N}{H}}
            -
            =
        ))
        -
        -(=\textcolor{O}{O})
        -\textcolor{N}{N}(-)
        -
    )
}}{\lstinline[breaklines=true]!Cn2c(=O)c1[nH]c(Cl)nc1n(C)c2=O!}
\end{lrbox}

\begin{tikzpicture}[style1/.style={rectangle, rounded corners,draw=black,thick,minimum size=10mm,font=\Large}, style2/.style={circle,thick, draw,minimum size=7mm,font=\Large}]
    \node[draw,style1, fill=yellow!50] (r) at (0,0) {RL};
    \node[draw, style1,right = of r, rounded corners, label={Final State (Generated Molecule)}, minimum width=70mm] (s) {\usebox{\MoleculeBox}};
    \draw[-latex, thick] (r.east) -- (s.west);
    \node[draw, style1, rounded corners, right = 2cm of s, fill=red!50] (p) {Penalty};
    
    \node[draw, style1, rounded corners, above = of p,  fill=orange!50] (i) {Int. Reward};
    \node[draw, style1, rounded corners, below = of p, fill=green!50] (e) {Ext. Reward};
    \draw[-latex] (s) -- (p);
    \draw let \p1 = ($(s.east)!0.5!(p.west)$), \p2 = (i.west) in 
     (\p1) -- (\x1,\y2) [-latex] (\x1,\y2) -- (\p2);
    \draw let \p1 = ($(s.east)!0.5!(p.west)$), \p2 = (e.west) in 
    (\p1) -- (\x1,\y2) [-latex,] (\x1,\y2) -- (\p2);
    \node[draw, style2, right = of e, fill=blue!50] (m) {};
    \draw[thick] (m.220) -- (m.45);
    \draw[thick] (m.135) -- (m.-45);
    \node[draw, style2, right = of m, fill=blue!50] (a) {};
    \draw[thick] (a.west) -- (a.east);
    \draw[thick] (a.south) -- (a.north);
    \draw let \p1 = (m.north), \p2 = (p.east) in
    (p.east)  --  (\x1,\y2)
    [-latex] (\x1,\y2) -- (m.north);
    \draw[-latex,] (e.east)  -- (m.west);
    \draw let \p1 = (a.north), \p2 = (i.east) in
    (i.east) -- (\x1,\y2)
    [-latex] (\x1,\y2) -- (\p1);
    \draw[-latex] (m.east) -- (a.west);
    \draw let \p1 = (e), \p2 = (a), \p3 = (r) in 
    (a.south) -- ($(\x2,\y1) - (0,1)$)
    ($(\x2,\y1) - (0,1)$) -- ($(\x3,\y1) - (0,1)$)
    [-latex,] ($(\x3,\y1) - (0,1)$) -- (r.south);
    \draw[dashed] ([shift={(-15pt,15pt)}]current bounding box.north west) rectangle ([shift={(15pt,-15pt)}]current bounding box.south east); 
    \node[draw, style1, rounded corners,left = of r, fill=red!50] (pr) {Prior};
    \draw[-latex, dashed] (pr.east) -- (r.west);
\end{tikzpicture}
    }
    \caption{The proposed diversity-aware RL framework for \textit{de novo} drug design utilizes extrinsic reward penalty and intrinsic reward to improve the diversity. The RL agent is initialized to the pre-trained prior. The RL agent generates molecules, e.g., in SMILES representation as shown here, and subsequently, the penalty and/or intrinsic reward is used to modify the extrinsic rewards. Each extrinsic reward is multiplied by the corresponding penalty term (equal to one if no penalty is used), while the intrinsic reward (equal to zero if no intrinsic reward is used) is added to the product. The modified rewards are observed by the RL agent and used to update its policy.}
    \label{fig:process}
\end{figure*}

The development of a novel pharmaceutical drug is a highly intricate process that can span up to a decade and incur costs exceeding US $\$1$ billion \citep{wouters2020estimated,dimasi2016innovation}. A key part of such effort involves the identification of novel drug candidates that exhibit the desired molecular properties \citep{hughes2011principles}. The success in identifying drug candidates primarily depends on selecting chemical starting points that surpass a certain threshold in bioactivity toward the desired target, known as hits. High-quality hits can substantially reduce the time required to identify a viable drug candidate and be the determining factor in the success of a drug discovery campaign \citep{quancard2023european}. Designing novel pharmaceutical molecules, or \textit{de novo} drug design, is extremely challenging given the estimated number of up to $10^{60}$ possible drug-like molecules \citep{reymond2015chemical}.

Recent advances in \textit{de novo} drug design utilize reinforcement learning (RL) to navigate this vast chemical space by fine-tuning a pre-trained generative model \citep{popova2018deep,gummesson2024utilizing,guo2024augmented,atance2022novo,liu2021drugex,olivecrona2017molecular}. Evaluations by \citet{gao2022sample} and \citet{thomas2022re} have demonstrated good performance when using RL to fine-tune a pre-trained recurrent neural network (RNN) \citep{rumelhart1985learning} to generate molecules encoded in the Simplified Molecular Input Line Entry System (SMILES) \citep{weininger1988smiles}. Also, this approach is widely adopted in real-world applications in drug discovery \citep{pitt2025real}.
However, RL-based \textit{de novo} drug design methods can easily become stuck in local optima, generating structurally similar molecules\--- a phenomenon known as \emph{mode collapse}. This is undesirable as it prevents the agent from discovering more diverse and potentially more promising local optima. To mitigate mode collapse, \citet{blaschke2020memory} introduced a count-based method that penalizes generated molecules based on their structure. When too many structurally similar molecules have been generated, the agent observes zero reward, instead of the actual extrinsic reward, for future generated molecules with the same structure. This is a popular way to avoid mode collapse for RL-based \textit{de novo} drug design \citep{thomas2022augmented, guo2024augmented, loeffler2024reinvent, gummesson2024utilizing}. 

Most work mainly focuses on avoiding mode collapse to find the most optimal solution. However, the quantitative structure-activity relationship (QSAR) models utilized for \emph{in silico} assessment of molecules introduce uncertainties and biases due to limited training data \citep{renz2019failure}. Thus, it is important to explore numerous modes of these models to increase the chance of identifying potential drug candidates. Also, the identified (local) optimal solution might not be optimal in terms of observed safety and therapeutic effectiveness in the body. Therefore, it is meaningful to generate a diverse set of molecules. Recent work by \citet{renz2024diverse} focuses on the generated molecules' diversity, finding superior performance of SMILES-based autoregressive models using RL to optimize the desired properties. They use a penalization method based on the work by \citet{blaschke2020memory} to enable diverse molecule generation.
As an alternative to penalizing the extrinsic reward, previous work in RL has shown that providing intrinsic motivation to the agent can enhance the exploration \citep{bellemare2016unifying,burda2018exploration, badia2020uplearningdirectedexploration}. Recent efforts by \citet{park2024molairmolecularreinforcementlearning} and \citet{wang2024exselfrl} show the potential of memory- and prediction-based intrinsic motivation approaches in \textit{de novo} drug design, demonstrating their capability to enhance the optimization of properties.   

The generation of diverse sets of molecules with high (extrinsic) rewards is crucial in the drug discovery process. A diverse molecular library increases the likelihood of identifying candidates with unique and favorable pharmacological profiles, thereby enhancing the overall efficiency and success rate of drug development pipelines. While most prior research has concentrated on generating individual molecules with high (extrinsic) rewards, our work shifts the focus toward the generation of diverse molecular entities by systematically investigating various intrinsic rewards and reward penalties. This approach aims to counteract mode collapse and promote the exploration of a broader chemical space. Intrinsic rewards, inspired by human-like curiosity, encourage the RL agent to explore less familiar areas of the chemical space; while reward penalties discourage the generation of structurally similar molecules. By employing these strategies, we aim to investigate further the robustness and applicability of RL-based de novo drug design. To our knowledge, this is the first work to comprehensively study the effect such methods have on the diversity of the generated molecules. By doing so, we provide a novel framework that not only seeks optimal solutions but also ensures a wide-ranging exploration of the potential chemical space of bespoke drug candidates. This could significantly enhance the drug discovery process by providing a more diverse and promising set of molecules for further experimental validation.


\section{Problem Formulation}
In this section, we introduce our framework for \textit{de novo} drug design. The problem is string-based molecule generation, by fine-tuning a pre-trained policy. Following previous work, we formulate the generative process as a reinforcement learning problem where the task is to fine-tune a pre-trained generative model \citep{olivecrona2017molecular}. 
An action corresponds to adding one token to the string representation of the molecule. $\mathcal{A}$ is the set of possible actions, including a start token $a^{\text{start}}$ and a stop token $a^{\text{stop}}$. The \textit{de novo} drug design problem can be modeled as a Markov decision process (MDP).  $a_t\in\mathcal{A}$ is the action taken at state $s_t$, the current state $s_t = a_{0:t-1}$ is defined as the sequence of performed actions up to round $t$, the initial action $a_0 = a^{\text{start}}$ is the start token. The transition probabilities $P(s_{t+1}|s_t,a_t) = \delta_{ s_t  \mdoubleplus a_t}$ are deterministic, where $P(\text{terminal}|s_t,a^{\text{stop}}) = 1$, $\mdoubleplus$ denotes the concatenation of two sequences and $\delta_z$ denotes the dirac distribution at $z$.
If action $a^{\text{stop}}$ is taken, the following state is terminal, stopping the current generation process and subsequently evaluating the generated molecule. The extrinsic reward is
\begin{equation}
    R(s_t,a_t) = R(a_{0:t}) =
    \begin{cases}
        r(s_{t+1}) & \text{if } a_t = a^{\text{stop}},\\
        0 & \text{otherwise.}
    \end{cases}
\end{equation} 
We let $T$ denote the round that a terminal state is visited, i.e., $a_{T-1} = a^{\text{stop}}$. 
The reward $r(s_{T}) \in [0,1]$ (only observable at a terminal state) measures the desired property, which we want to optimize, of molecule $A = a_{1:T-2}$. Note that in practice, the string between the start and stop tokens encodes a molecule such that $a_{1:T-2}$ is equivalent to $a_{0:T-1}$ during evaluation. The objective is to fine-tune a policy $\pi_\theta$, parameterized by $\theta$, to generate a structurally diverse set of molecules optimizing the property score $r(\cdot)$.

In practice, at each step $i$ of the generative process, $B$ full trajectories (until reaching a terminal state) are rolled out, to obtain a batch $\mathcal{B}$ of generated molecules. Also, the diversity-aware reward $\hat{R}(A)$ (see \cref{sec:intrinsic_rewards_penalties}) for each molecule $A\in\mathcal{B}$ is observed by the agent and subsequently used for fine-tuning. The diversity-aware reward $\hat{R}(A)$ is computed using the penalty $f(A)$ and/or intrinsic reward $R_I$ (depending on which reward function is used). \Cref{alg:problem} illustrates our diversity-aware RL framework. 
\begin{algorithm}
\caption{Diversity-Aware RL framework}\label{alg:problem} 
\begin{algorithmic}[1] 
\State \textbf{input:} $I,B,\theta_{\text{prior}},h$
\State $\mathcal{M} \gets \emptyset$ \Comment{Initialize memory}
\State $\theta \gets \theta_{\text{prior}}$ \Comment{The pre-trained policy is fine-tuned}
 \For{i=1,\dots,I} \Comment{Generative steps}
 \State $L(\theta) \gets 0$ 
 \State $\mathcal{B} \gets \emptyset$
    \For{b=1,\dots,B} \Comment{Generate batch of molecules}
        \State $t\gets 0$
        \State $a_t \gets a^{(\text{start})}$ \Comment{Start token is always initial action}
        \State $s_{t+1} \gets a_{t}$
        \While{$s_{t+1}$ is not terminal}
            \State $t\gets t+1$
            \State $a_t \sim \pi_\theta(s_t)$
            \State $s_{t+1} \gets a_{0:t}$
        \EndWhile
        \State $\mathcal{B} \gets \mathcal{B} \cup s_{t+1}$
        \State Observe property score $r(s_{t+1})$
        \If{$r(s_{t+1}) \geq h$}
            \State $\mathcal{M} \gets \mathcal{M} \cup \{s_{t+1}\}$
        \EndIf  
        \State Compute and store penalty $f(s_{t+1})$
    \EndFor
    \For{$A\in\mathcal{B}$} 
        \State Compute intrinsic reward $R_I(A)$ 
        \State Compute diversity-aware reward $\hat{R}(A)$
        \State Compute loss $L_A (\theta)$ wrt $\hat{R}(A)$ 
        \State $L(\theta) \gets L(\theta) + L_A (\theta)$
    \EndFor
    \State Update $\theta$ by one gradient step minimizing $L(\theta)$ 
\EndFor

 \State \textbf{output:} $\mathcal{M}$
\end{algorithmic} 
\end{algorithm}

\section{Diversity-Aware Reward Functions}
\label{sec:intrinsic_rewards_penalties}
In this section, we define the diversity-aware reward functions examined in this study. We investigate two approaches to encourage diversity among generated molecules by modifying the extrinsic reward: (1) penalize the extrinsic reward, and (2) provide intrinsic reward (intrinsic motivation). Moreover, we also investigate the combination of these approaches by integrating two intrinsic reward approaches with a penalty function on the extrinsic reward. Given an extrinsic reward $R(A)$, the agent will receive a reward signal at the end of the generation sequence in the form of  
\begin{equation}
    \hat{R}(A) = f(A) \times R(A) + R_I(A),
\end{equation}
for every generated molecule $A$. Hence, we impose reward shaping \citep{laud2004theory} where the reward observed by the agent is a linear function of the non-linear extrinsic reward, as illustrated in \cref{fig:process}. The penalty defines the importance of the extrinsic reward for each molecule to determine the exploitation rate adaptively. Sufficient exploitation is necessary to find high-quality solutions. In contrast, the intrinsic reward provides an additive bonus to encourage the agent to continually explore (independent of a molecule's extrinsic reward). We suggest several novel domain-specific penalties and intrinsic rewards, and, to the best of our knowledge, this domain-specific combination is novel. Such a linear combination avoids adding unnecessary complexity to the reward objective, but is necessarily not optimal. We want to optimize a complex extrinsic reward function while enforcing continuous exploration of a large solution space. The aim is to find a diverse set of high-quality solutions.

\subsection{Extrinsic Reward Penalty}
We propose and examine five different functions to penalize the extrinsic reward by discretely or continuously decreasing it based on the number of previously generated structurally similar molecules. These are based on binary, error, linear, sigmoid, or hyperbolic tan functions. To the best of our knowledge, utilizing error and hyperbolic tangent functions is novel for our application. Below we define all penalty functions for clarity. 

Non-binary functions will provide an incremental change of the (extrinsic) reward signal over time. Therefore, it is potentially more informative and can incrementally incentive the agent to find new local optima, while still exploiting the current local optima. On the other hand, a binary function implies a sharp change of the reward function, where the agent quickly needs to find new optima.

\paragraph{Identical Molecular Scaffold Penalty (IMS)} The IMS penalty was first introduced by \citet{blaschke2020memory} and has thereafter been used in several works, e.g., \citep{loeffler2024reinvent,guo2024augmented}. It is based on molecular scaffolds, which is one of the most important and commonly used concepts in medicinal chemistry. The IMS penalty uses the \emph{molecular scaffold} defined by \citet{bemis1996properties}, which is obtained by removing all side chains (or R groups).
In this work, we also study the Topological scaffold, which is obtained from the molecular scaffold by converting all atom types into carbon atoms and all bonds into single bonds. Note that in this work we use the molecular scaffold since it is less general and has demonstrated good performance in earlier works \citep{blaschke2020memory,guo2024augmented,thomas2022augmented}. The Topological scaffold is therefore exclusively applied to assess the molecules' diversity and is not incorporated into any penalty or intrinsic reward method defined hereafter. 

Let us define the reward function $\hat{R}_{\text{IMS}}(A)$ for the IMS penalty. For each generated molecule $A$ with a reward of at least $h$, its molecular scaffold $S_A$ is computed and put in memory. A molecule fulfilling the (extrinsic) reward threshold $h$ is commonly known as a predicted active molecule, denoted simply as \emph{active}. If $m$ molecules with the same scaffold have been generated, future molecules of the same scaffold are given a reward of $0$ to avoid this scaffold. Given a generated molecule $A$ with an extrinsic reward of at least $h$ and its corresponding molecular scaffold $S_A$, the reward function of the IMS penalty method is then defined by
\begin{equation}
    \hat{R}_{\text{IMS}}(A) = 
        \begin{cases}
            0 & \text{if $R(A) \geq h$ and $N[S_A] \geq m$},\\
            R(A) & \text{otherwise},\\
        \end{cases}
\end{equation}
where $S_A$ is the molecular scaffold of molecule $A$ and $N[S]$ is the number of molecules with molecular scaffold $S$ in memory, i.e., with an extrinsic reward of at least $h$.
If a molecule $A$ corresponds to an extrinsic reward smaller than $h$, the extrinsic reward is provided to the agent without any modification. Hence, only predicted active molecules are penalized.

\paragraph{Error Function Identical Molecular Scaffold Penalty (ErfIMS)} The Error Function Identical molecular scaffold Penalty is a soft (non-binary) version of the IMS penalty method. It uses the error function to incrementally decrease extrinsic rewards based on the number of molecules in the molecular scaffold
\begin{equation}
    f_{\text{erf}}(A) = \left(1 + \text{erf}\left(\frac{\sqrt{\pi}}{m}\right) - \text{erf}\left( \frac{\sqrt{\pi}\times N[S_A]}{m}\right)\right),
\end{equation}
where $\text{erf}(x) = \frac{2}{\sqrt{\pi}}\int_0^x e^{-t^2} dt$ is the error function.
 Given the threshold $h$ for the extrinsic reward $R(\cdot)$, the reward function for a molecule $A$ is defined by
\begin{equation}
    \hat{R}_{\text{ErfIMS}}(A) = 
        \begin{cases}
           R(A)\cdot  f_{\text{erf}}(A) & \text{if $R(A) \geq h$},\\
           R(A) & \text{otherwise}.
        \end{cases}
\end{equation}

\paragraph{Linear Identical molecular scaffold Penalty (LinIMS)}
The linear identical molecular scaffold penalty linearly reduces the extrinsic score based on the number of generated molecules in memory with the same molecular scaffold. We define the linear penalty function by
\begin{equation}
    f_{\text{linear}}(A) = \left(1-\frac{N[S_A]}{m}\right).
\end{equation}
The reward function of a molecule $A$ is defined by, given the threshold $h$ for the extrinsic reward $R(\cdot)$,
\begin{equation}
    \hat{R}_{\text{LinIMS}}(A) = 
        \begin{cases}
           \left[R(A)\cdot  f_{\text{linear}}(A)\right]^+ &  \text{if $R(A) \geq h$},\\
           R(A) & \text{otherwise}, \\
        \end{cases}
\end{equation}
where $[\cdot]^+$ denotes the positive part of a function. This is equivalent to the linear penalty proposed by \citet{blaschke2020memory}.

\paragraph{Sigmoid Identical Molecular Scaffold Penalty (SigIMS)}
The sigmoid identical molecular scaffold penalty uses a sigmoid function to gradually reduce the extrinsic reward based on the number of molecules in memory with the same scaffold. A molecule is put in memory if it has an extrinsic reward of at least $h$. Given a molecule $A$, the sigmoid function in this work is defined as 
\begin{equation}
    f_{\sigma} (A) = 1- \frac{1}{1 + e^{- \left(  \frac{\frac{N\left[S_A\right]}{m}\cdot 2-1}{0.15}\right)}}.
\end{equation}
 This is equivalent to the sigmoid penalty function proposed by \citet{blaschke2020memory} and we therefore use the same parameters. Given a molecule $A$, we define the reward function defined as follows
\begin{equation}
    \hat{R}_{\text{SigIMS}}(A) = 
        \begin{cases}
           R(A)\cdot f_{\sigma} (A)  & \text{if $R(A) \geq h$},\\
           R(A) & \text{otherwise}. \\
        \end{cases}
\end{equation}

\paragraph{Tanh Identical Molecular Scaffold Penalty (TanhIMS)}
The tanh identical molecular scaffold penalty utilizes the hyperbolic tangent function to incrementally decrease the extrinsic reward based on the number of molecules generated with the same molecular scaffold, up to and including the current step (i.e., those stored in memory). For a molecule $A$, the following hyperbolic tangent function is used to incrementally penalize the extrinsic reward
\begin{equation}
    f_{\tanh} \left(A\right) = 1- \tanh \left( c_{\text{tanh}}\cdot \frac{N[S_A]-1}{m}\right). 
\end{equation}
In the following experiments we use $c_{\text{tanh}}=3$ since this factor implies $f_{\tanh}\approx 0$ around $N[S_A] = 25$ for a bucket size $m=25$.
For a molecule $A$, we define the reward function for the TanhIMS penalty as follows
\begin{equation}
    \hat{R}_{\text{TanhIMS}}(A) = 
        \begin{cases}
           R(A)\cdot f_{\tanh}(A) & \text{if $R(A) \geq h$},\\
           R(A) & \text{otherwise}. \\
        \end{cases}
\end{equation}

\subsection{Intrinsic Reward}
We explore eight methods to provide intrinsic reward to the agent, namely diverse actives (DA), minimum distance (MinDis), mean distance (MeanDis), minimum distance to random coreset (MinDisR), mean distance to random coreset (MeanDisR), KL-UCB, random network distillation (RND) and information (Inf). To the best of our knowledge, all methods, except for RND, are novel in the context of \textit{de novo} drug design.

\paragraph{Diverse Actives (DA)} We define the diverse actives intrinsic reward based on the diverse hits metric by \citet{renz2024diverse}, which is based on \#Circles metric proposed by \citet{xie2023much}. Given a set of possible centers, the \#Circles metric counts the number of non-overlapping circles with equivalent radius in the distance metric space. An \emph{active} molecule is defined as a molecule with a reward of at least $h$. Following the work by \citet{renz2024diverse} but using the terminology of predicted active molecules rather than hit molecules, we define the number of \emph{diverse actives} for distance threshold $D$ by
\begin{equation}
\label{eq:diverse_active}
\resizebox{0.89\hsize}{!}{%
    $\mu\left(\mathcal{H}; D\right) = \max_{\mathcal{C}\in \mathcal{P}\left(\mathcal{H} \right)} |\mathcal{C}|\text{ s.t. } \forall x \neq y \in \mathcal{C}: d(x,y) \geq D,$ %
    }
\end{equation}
where $\mathcal{H}$ is a set of predicted active molecules, $\mathcal{P}$ is the power set, $d(x,y)$ is the distance between molecules $x$ and $y$. Note that there is a substantial difference between a set of actives and a set of diverse actives. Determining the number of diverse actives is analogous to determining the packing number of the set $\mathcal{H}$ in the distance metric space \citep{renz2024diverse}. Let $\Delta_{\mu}$ be the difference in the number of diverse actives, between two sets $\mathcal{H}$ and $\widetilde{\mathcal{H}}$ of active molecules, defined by
\begin{equation}
    \Delta_{\mu}\left(\mathcal{H}, \widetilde{\mathcal{H}};D\right) = \mu\left(\mathcal{H};D\right)-\mu\left(\widetilde{\mathcal{H}};D\right).
\end{equation}
Moreover, let $\mathcal{H}_i$ be the batch of generated actives in the current generative step $i$, $\mathcal{C}_{i-1}$ the set of previously generated diverse actives and $\mathcal{C}_{i} = \mathcal{C}_{i-1}\cup \mathcal{H}_i$. We define the reward function using diverse actives as an intrinsic reward by
\begin{equation}
    \hat{R}_{\text{DA}}(A) = 
        \begin{cases}
           R(A) + \Delta_{\mu}\left(\mathcal{C}_{i}, \mathcal{C}_{i-1};D\right) & \text{if $A\in \mathcal{H}_i$},\\
           R(A) & \text{otherwise}. \\
        \end{cases}
\end{equation}
Note that the intrinsic reward $\Delta_{\mu}\left(\mathcal{C}_{i-1}\cup \mathcal{H}_i, \mathcal{C}_{i-1};D\right) $ is sparse since a new batch does not necessarily increase the number of non-overlapping circles. On the other hand, the intrinsic reward can be substantially larger than the extrinsic reward $R(A)\in[0,1]$, providing strong intrinsic motivation towards a specific area. 

\paragraph{Minimum Distance (MinDis)} Minimum distance is a distance-based intrinsic reward. A bonus reward is given based on the minimum distance to previously generated diverse actives (see definition of diverse actives above). 
Let $\mathcal{H}_i$ be a batch of generated actives in the current generative step $i$ and $\mathcal{C}_{i-1}$ be a set of previously generated diverse actives. Then the reward function of MinDis for a molecule $A$ and reward threshold $h$ (of predicted active molecules) is defined as follows
\begin{equation}
\resizebox{0.89\hsize}{!}{%
        $\hat{R}_{\text{MinDis}}(A) = 
            \begin{cases}
               R\left(A\right) + \displaystyle \min_{\tilde{A}\in \widetilde{\mathcal{C}}_{i-1}}  d\left(A,\tilde{A}\right)  & \text{if $R\left(A\right) \geq h$},\\
               R\left(A\right) & \text{otherwise}, \\
            \end{cases}$%
            }
\end{equation}
where $\widetilde{\mathcal{C}}_{i-1} \coloneq \mathcal{C}_{i-1} \cup \left(\mathcal{H}_i\setminus \{A\}\right)$, and $d(x,y)$ is a distance metric between molecules $x$ and $y$. In this work, we use the distance metric based on the Jaccard index \citep{jaccard1912distribution}, also known as the Tanimoto distance, widely used to measure chemical (dis-)similarity.

\paragraph{Mean Distance (MeanDis)} Mean distance is also a distance-based intrinsic reward, but where the intrinsic reward is defined as the mean dissimilarity (distance) to previously generated diverse actives and the current batch of actives. 
Let $\mathcal{H}_i$ be a batch of generated actives in the current generative step $i$ and $\mathcal{C}_{i-1}$ be a set of previously generated diverse actives (see definition above of diverse actives). We then define the reward function of MeanDis of molecule $A$ by
\begin{equation}
    \hat{R}_{\text{MeanDis}}(A) = 
        \begin{cases}
           R(A) + \bar{d} (A; \widetilde{\mathcal{C}}_{i-1})  & \text{if $R(A) \geq h$},\\
           R(A) & \text{otherwise}, \\
        \end{cases}
\end{equation}
where $\bar{d} (A; \widetilde{\mathcal{C}}_{i-1}) = \frac{\sum_{\tilde{A} \in \widetilde{\mathcal{C}}_{i-1}} d(A, \tilde{A})}{|\widetilde{\mathcal{C}}_{i-1}|}$.

\paragraph{Minimum Distance to Random Coreset (MinDisR)} Minimum distance to random coreset is a distance-based intrinsic reward similar to MinDis. The difference is that MinDisR is based on the distance between the actives from the current generative step and a random set of previously generated actives. 
Given the set $\mathcal{H}_i$ of actives generated in the current generative step $i$, a molecule $A\in \mathcal{H}_i$ generated in the current step, and a (uniform) random set $\mathcal{X}$ of previously generated actives, the reward function is defined by
\begin{equation}
\resizebox{0.89\hsize}{!}{%
    $\hat{R}_{\text{MinDisR}}(A) = 
        \begin{cases}
           R(A) + \displaystyle \min_{\tilde{A}\in \widetilde{\mathcal{X}} } d\left(A,\tilde{A}\right)  & \text{if $R(A) \geq h$}\\
           R(A) & \text{otherwise}, \\
        \end{cases}$%
        }
\end{equation}
where $\widetilde{\mathcal{X}} \coloneq \mathcal{X} \cup \left(\mathcal{H}_i \setminus \{A\} \right)$ and $d(x,y)$ is the distances between molecules $x$ and $y$. In this work, $\mathcal{X}$ consists of 5000 randomly sampled actives, uniformly sampled without replacement from the set of previously generated actives $\mathcal{H}_{i-1}$. If 5000 actives have not been generated at generative step $i$, all previously generated actives are used.

\paragraph{Mean Distance to Random Coreset (MeanDisR)} Mean distance to random coreset is an intrinsic reward given by the mean distance to a random coreset of actives. Given a set $\mathcal{H}_i$ of actives generated in the current generative step $i$, a molecule $A$ generated in the current generative step and a random set $\mathcal{X}$ of previously generated actives, we define the reward function of MeanDisR as 
\begin{equation}
    \hat{R}_{\text{MeanDisR}}(A) = 
        \begin{cases}
           R(A) +  \bar{d} (A; \tilde{\mathcal{X}})  & \text{if $R(A)\geq h$}\\
           R(A) & \text{otherwise}. \\
        \end{cases}
\end{equation}

\paragraph{KL-UCB} The KL-UCB intrinsic reward is based on the KL-UCB algorithm by \citet{pmlr-v19-garivier11a} for the multi-armed bandit problem. It defines an improved upper confidence bound to handle the trade-off between exploration and exploitation in the multi-armed bandit problem. In our study, this trade-off is crucial as the agent must determine the optimal balance between exploiting and exploring various molecular structures. We compute the KL-UCB intrinsic reward for a molecule $A$ by
\begin{equation}
\resizebox{0.89\hsize}{!}{%
    $\begin{split}
    R_I^{\text{UCB}}(A) &= \max \left\{ q \in [0,1] : N[S_A] \text{KL}\left(\frac{\Sigma[S_A]}{N[S_A]},q\right) \right.\\
    &\left.\leq \log (n) + c \log (log(n))\right\},
    \end{split}$%
    }
\end{equation}
where $n$ is the total number of generated molecules up to and including the current generative step $i$,  $\text{KL}\left(p,q\right) = p \log \frac{p}{q} + (1-p) \log \frac{1-p}{1-q}$ is the Bernoulli Kullback-Leibler divergence and $c=0$ is used for optimal performance in practice \citep{pmlr-v19-garivier11a}. Moreover, $S_A$ is the scaffold of molecule $A$, $\Sigma[S]$ is the sum of rewards of actives with scaffold $S$ in memory and $N[S]$ is the total number of actives with scaffold $S$ in memory. 
Given a molecule $A$, extrinsic reward $R(A)$ and reward threshold $h$, we define the reward function of KL-UCB by
\begin{equation}
    \hat{R}_{\text{KL-UCB}}(A) = 
        \begin{cases}
           R_I^{\text{UCB}}(A)  & \text{if $R(A) \geq h$},\\
           R(A) & \text{otherwise}. \\
        \end{cases}
\end{equation}
This implies that actives are given a reward corresponding to the upper confidence bound of the mean extrinsic reward of the actives with the same scaffold.

\paragraph{Random Network Distillation (RND)} Random network distillation \citep{burda2018exploration} is an exploration technique in reinforcement learning that provides an intrinsic reward based on the prediction error of a neural network. Specifically, it employs a fixed, randomly initialized neural network $f$ and a predictive neural network $\hat{f}_{\phi}$ trained to mimic the outputs of the fixed network. The intrinsic reward is derived from the prediction error between these two networks. This error serves as a measure of novelty, incentivizing the RL agent to explore less familiar regions of the parameter space, and potentially enhancing the exploration of less familiar regions of the chemical space. We adapt RND as an intrinsic reward for generated active molecules, i.e., molecules with an extrinsic reward of at least $h$. In this work, $f$ and $\hat{f}_{\phi}$ have identical architecture as the pre-trained policy (see \cref{sec:experiments}). We let the predictive network $\hat{f}_{\phi}$ be initialized to this pre-trained policy. For a molecule $A$, we define 
\begin{align}
    f(A) &= \sum_{t=1}^{T-2} \log \pi_f (a_t|s_t),\\
    \hat{f}_\phi(A) &= \sum_{t=1}^{T-2} \log \pi_\phi (a_t|s_t),
\end{align}
where $\pi_f (a_t|s_t)$ and $\pi_\phi (a_t|s_t)$ are the policies induced by the fixed and predictive network, respectively. Moreover, let $\Delta_{\hat{f}}(A;\phi)$ be the squared norm of the difference between these networks for a molecule $A$, defined by
\begin{equation}
    \Delta_{\hat{f}}(A;\phi) = \|\hat{f}_\phi (A) - f(A)\|^2,
\end{equation}
where $\phi$ is the weights of the prediction network $\hat{f}$ of the current generative step.

Let $\phi$ be the weights of the predictive network $\hat{f}$ up to the current generative step, we define the reward function of a molecule $A$ by
\begin{equation}
    \hat{R}_{\text{RND}}(A) = 
        \begin{cases}
           R(A) + \Delta_{\hat{f}}(A;\phi)  & \text{if $R(A) > h$},\\
           R(A) & \text{otherwise}. \\
        \end{cases}
\end{equation}
Since $R(A) \in [0,1]$, we rescale the prediction error over the batch of active molecules generated in the current generative step $i$ by 
\begin{equation}
     \tilde{\Delta}_{\hat{f}}(A;\phi) = \frac{\Delta_{\hat{f}}(A;\phi) - \displaystyle \min_{\tilde{A}\in \mathcal{H}_{i} } \Delta_{\hat{f}}(\tilde{A};\phi)}{\displaystyle \max_{\tilde{A}\in \mathcal{H}_{i} } \Delta_{\hat{f}}(\tilde{A};\phi) -  \displaystyle\min_{\tilde{A}\in \mathcal{H}_{i} } \Delta_{\hat{f}}(\tilde{A};\phi)}.
\end{equation}

\paragraph{Information (Inf)} We define an information-inspired intrinsic reward function based on the number of actives in each scaffold and scaffolds generated up to and including the current generative step $i$. Let $A$ be a molecule,  $S_A$ its scaffold, $N[S]$ the number of active molecules with scaffold $S$ in memory, and $\mathcal{S}$ the set of unique molecular scaffolds in memory up to (including) current generative step $i$. We define the the \emph{scaffold} (pseudo-)\emph{probability} of molecule $A$ by
\begin{equation}
    \tilde{\mathbb{P}}_{\text{scaff}} \left(A\right) = \frac{N[S_A]}{|\mathcal{S}|}.
\end{equation} 
We use this scaffold probability to define the \emph{scaffold information} by
\begin{equation}
   \text{I}_{\text{scaff}} (A) =  -\log \left(\tilde{\mathbb{P}}_{\text{scaff}} \left(A\right)\right).
\end{equation}
Given a set of active molecules $\mathcal{H}_i$ generated at the current generative step $i$, where $A \in \mathcal{H}_i$, the normalized scaffold information is defined by
\begin{equation}
    R^{\text{Inf}}_{I} \left(A; \mathcal{H}_i\right) = \frac{ \text{I}_{\text{scaff}} \left(A\right)  - \displaystyle \min_{\tilde{A}\in \mathcal{H}_i}  \text{I}_{\text{scaff}} \left(\tilde{A}\right)}{\displaystyle \max_{\tilde{A}\in \mathcal{H}_i}  \text{I}_{\text{scaff}} \left(\tilde{A}\right) - \displaystyle \min_{\tilde{A}\in \mathcal{H}_i}  \text{I}_{\text{scaff}} \left(\tilde{A}\right)}.
\end{equation}
In practice, we only normalize if $|\mathcal{H}_i| > 2$. Using the scaffold information to define the information-based intrinsic reward, we define the reward function by
\begin{equation}
    \hat{R}_{\text{Inf}}\left(A\right) = 
        \begin{cases}
           R\left(A\right) +  R^{\text{Inf}}_{I} \left(A; \mathcal{H}_i\right) & \text{if $R\left(A\right) \geq h$},\\
           R\left(A\right) & \text{otherwise}. \\
        \end{cases}
\end{equation}

\subsection{Combining Penalty and Intrinsic Reward}
We investigate two combinations of intrinsic reward and extrinsic reward penalty. To the best of our knowledge, combining these approaches is novel.

\paragraph{Tanh Random Network Distillation (TanhRND)} We define a soft version of random network distillation by combining RND and TanhIMS. The extrinsic reward is penalized as defined by TanhIMS and an (non-penalized) intrinsic reward based on RND is provided to the agent. We define the TanhRND reward function by 
\begin{equation}
\resizebox{0.89\hsize}{!}{%
    $\hat{R}_{\text{TanhRND}}\left(A\right) = 
        \begin{cases}
            R_{\tanh}\left(A\right) + \tilde{\Delta}_{\hat{f}}(A;\phi)  & \text{if $R\left(A\right) \geq h$},\\
           R\left(A\right) & \text{otherwise}, \\
        \end{cases}$%
        }
\end{equation}
where $R_{\tanh}\left(A\right) = f_{\tanh}\left(A\right)\cdot R\left(A\right)$.

\paragraph{Tanh Information (TanhInf)} We also propose and examine a reward function combining (extrinsic) reward penalty and information-based intrinsic reward. We use TanhIMS to penalize the extrinsic reward and Inf to provide a (non-penalized) intrinsic reward. For a molecule $A$, we define the TanhInf reward function by 
\begin{equation}
\resizebox{0.89\hsize}{!}{%
    $\hat{R}_{\text{TanhInf}}\left(A\right) = 
        \begin{cases}
           R_{\tanh}\left(A\right) + \tilde{\text{I}}_{\text{scaff}} \left(A; \mathcal{H}_i\right)  & \text{if $R\left( A\right) \geq h$},\\
           R\left(A\right) & \text{otherwise}, \\
        \end{cases}$%
        }
\end{equation}
where $h$ is the reward threshold, $R\left(\cdot\right)$ is the extrinsic reward, $\mathcal{H}_i$ is the set of active molecules generated in the current generative step.


\section{Experimental Evaluation}
\label{sec:experiments}
We now describe experiments designed to examine the efficacy of our diversity-aware reward functions.

\subsection{Experimental Setup}
We run experiments on three extrinsic reward functions, namely the
c-Jun N-terminal Kinases-3 (JNK3), Glycogen Synthase Kinase 3 Beta (GSK3$\beta$) and Dopamine Receptor D2 (DRD2) oracle provided by Therapeutics Data Commons \citep{Velez-Arce2024tdc}. These are well-established molecule binary bioactivity label optimization tasks. 
To compute an extrinsic reward in $[0,1]$, each oracle utilizes a random forest classifier trained on data from the ExCAPE-DB dataset \citep{sun2017excape} using extended-connectivity fingerprints with radius 3 \citep{rogers2010extended}. 
These oracles only provide rewards to valid molecules and, therefore, we assign invalid molecules an extrinsic reward of $-1$ to distinguish them from the penalized molecules. Additionally, previously generated (predicted) active molecules are each assigned zero reward.

For distance-based intrinsic rewards, the Jaccard distance is computed based on Morgan fingerprints \citep{rogers2010extended} computed by RDKit \citep{landrum2006rdkit}, with a radius of 2 and a size of 2048 bits. The distance threshold for the diverse actives-based approaches is fixed to $D=0.7$, as suggested by \citet{renz2024diverse} since there is a significant decrease in the probability of similar bioactives beyond this threshold \citep{10.12688/f1000research.8357.2}. Moreover, scaffold-based diversity-aware reward functions (see \cref{sec:intrinsic_rewards_penalties}) utilize molecular scaffolds and a bucket size of $m = 25$. The reward that the agents see is only modified for molecules with an extrinsic reward of at least $0.5$, i.e.,  we define predicted active molecules as molecules reaching an (extrinsic) reward of at least $h = 0.5$.

The molecular generative model builds directly on REINVENT \citep{segler2018generating, olivecrona2017molecular,blaschke2020reinvent,loeffler2024reinvent} and consists of a long short-term memory (LSTM) network \citep{hochreiter1997long} using SMILES to represent molecules as text strings. REINVENT  utilizes an on-policy RL algorithm optimizing the policy $\pi_\theta$ to generate molecules with higher reward.  
 The algorithm is based on the \emph{augmented log-likelihood} defined by 
\begin{equation}
    \log \pi_{\theta_{\text{aug}}}(A) \coloneq \sum_{t=1}^{T-2}\log \pi_{\theta_{\text{prior}}}\left(a_t|s_{t}\right) + \sigma R(A),
\end{equation}
where $A=a_{1:T-2}$ is a generated molecule, $\sigma$ is a scalar value, $\pi_{\theta_{\text{prior}}}$ is the (fixed) prior policy. We use the pre-trained policy by \citet{blaschke2020reinvent} as the prior policy. It is pre-trained on the ChEMBL database \citep{gaulton2017chembl} to generate drug-like bioactive molecules. The action space $\mathcal{A}$ consists of 34 tokens including start and stop tokens, i.e., $|\mathcal{A}| = 34$. The policy $\pi_\theta$ is optimized by minimizing the squared difference between the augmented log-likelihood and policy likelihood given a sampled batch $\mathcal{B}$ of SMILES
\begin{equation}
\label{eq:loss}
\begin{split}
    L(\theta) = \frac{1}{|\mathcal{B}|}  \sum_{a_{1:T-2}\in \mathcal{B}} \left(\log \pi_{\theta_{\text{aug}}}(a_{1:T-2})\right. \\
    \left. - \sum_{t=1}^{T-2}\log \pi_{\theta}(a_t|s_t)\right)^2.
    \end{split}%
\end{equation}
Previous work has shown that minimizing this loss function is equivalent to maximizing the expected return, as for policy gradient algorithms \citep{guo2024augmented}. Evaluations by both \citet{gao2022sample} and \citet{thomas2022re} have concluded good performance compared to both RL-based and non-RL-based approaches for \textit{de novo} drug design.
The generative process has a budget of $I = 2000$ generative steps, where a batch of $|\mathcal{B}|=128$ molecules is generated in each step. Each experiment is evaluated by 20 independent runs of the RL fine-tuning process.  

\subsection{Comparison of Diversity-Aware Reward Functions}
For both the GSK3$\beta$- JNK3-and DRD2-based extrinsic rewards, we evaluate the quality by extrinsic reward per generative step, and diversity by the number of molecular scaffolds, topological scaffolds and diverse actives after $I=2000$ generative steps.
\paragraph{GSK3$\beta$}
\begin{figure}[h]
     \centering
     \begin{subfigure}[b]{0.23\textwidth}
         \centering
         \includegraphics[width=\textwidth]{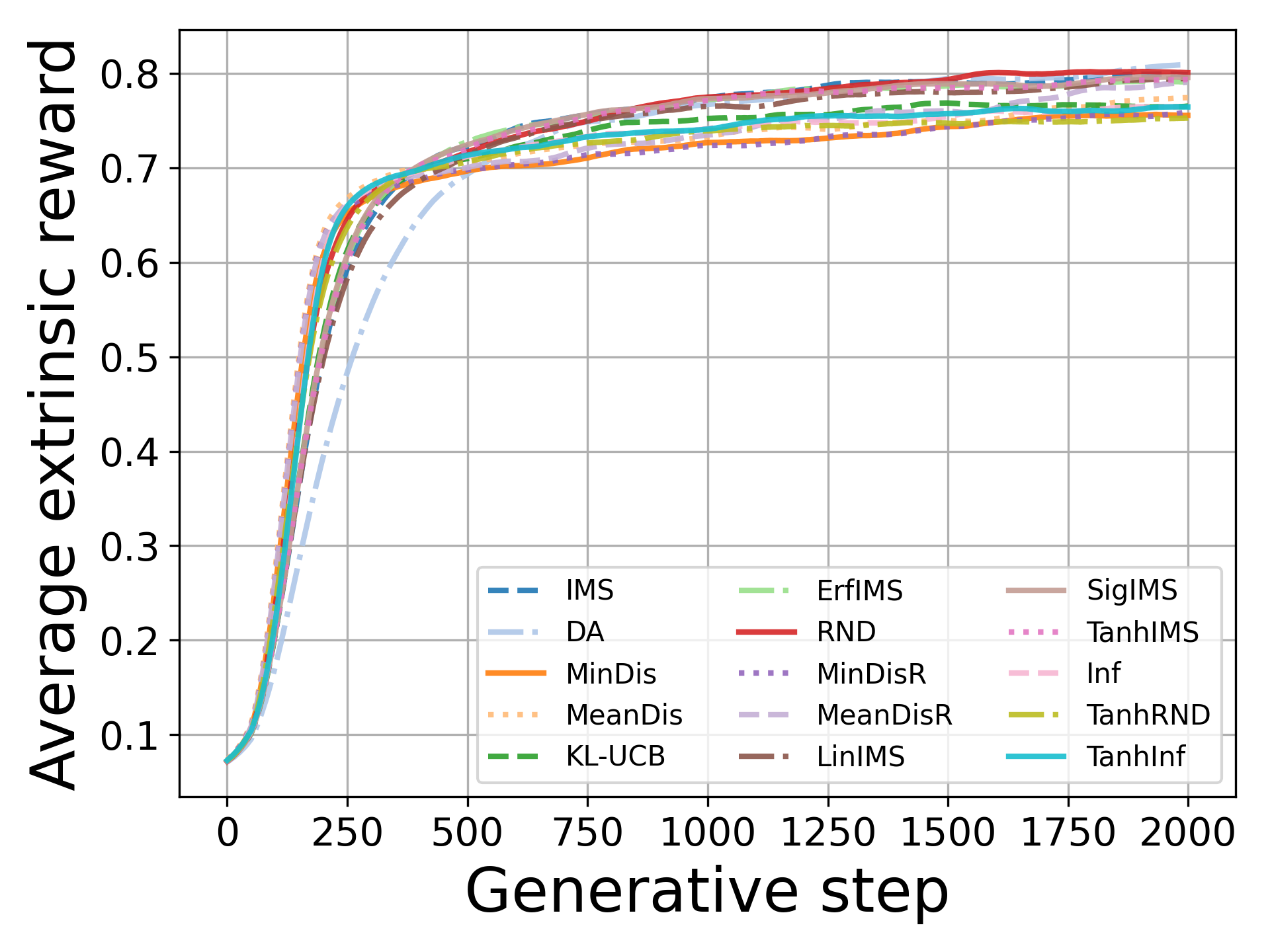}
         \caption{Extrinsic reward}
         \label{fig:gsk3b_reward}
     \end{subfigure}
     \hfill
     \begin{subfigure}[b]{0.23\textwidth}
         \centering
         \includegraphics[width=\textwidth]{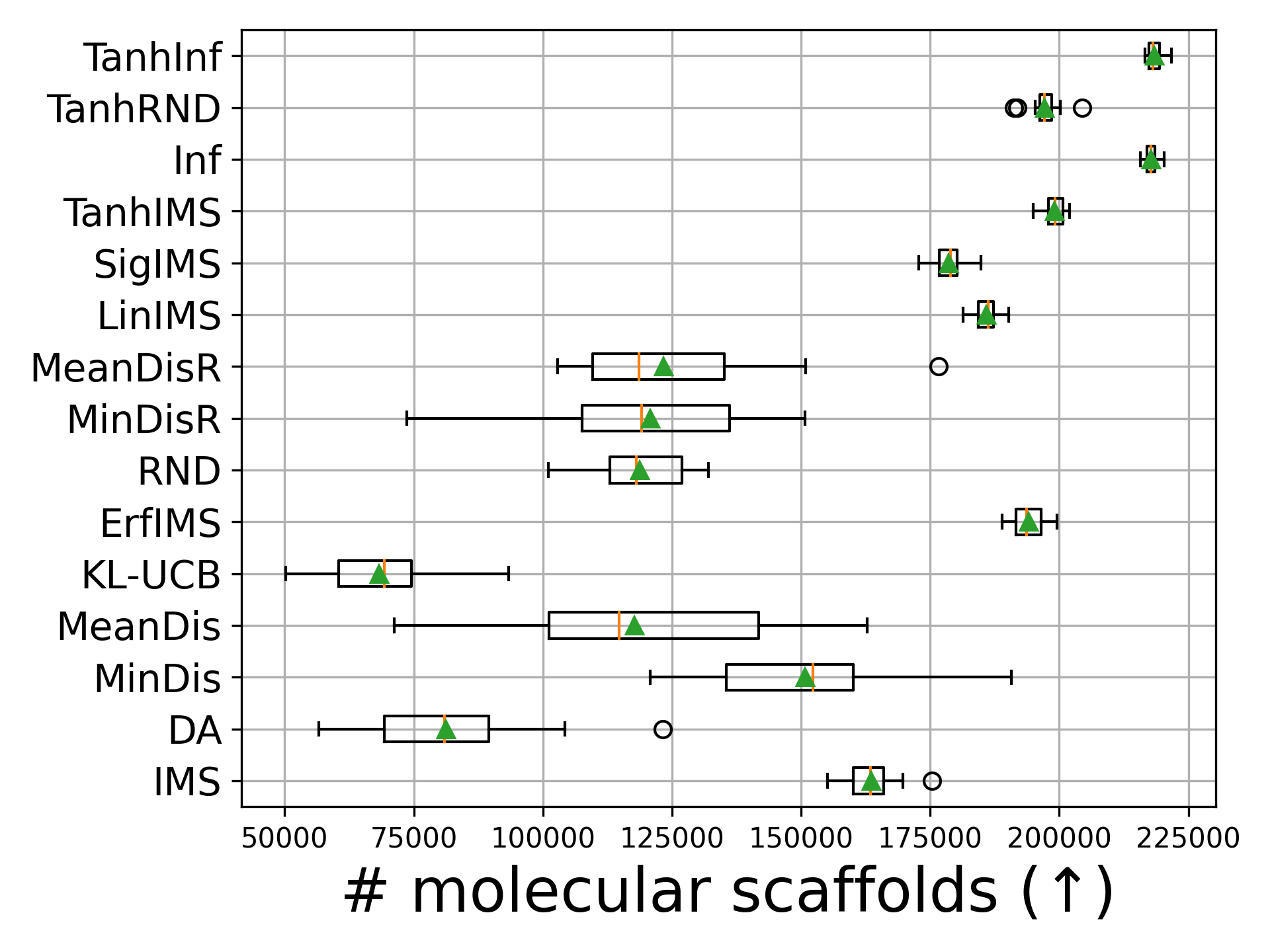}
         \caption{Molecular scaffold}
         \label{fig:gsk3b_chemical}
     \end{subfigure}
     \begin{subfigure}[b]{0.23\textwidth}
         \centering
         \includegraphics[width=\textwidth]{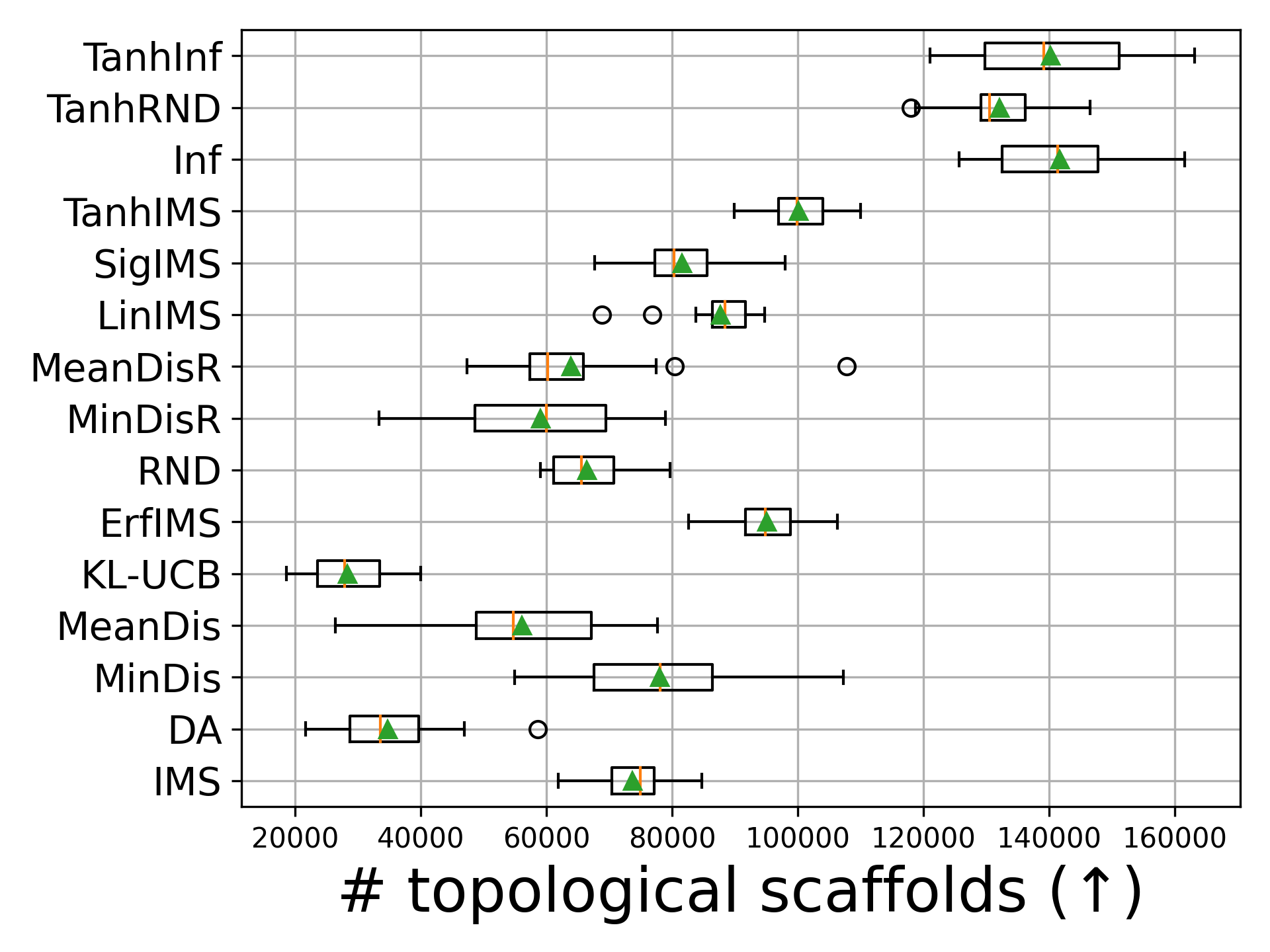}
         \caption{Topological scaffold}
         \label{fig:gsk3b_topo}
     \end{subfigure}
     \hfill
     \begin{subfigure}[b]{0.23\textwidth}
         \centering
         \includegraphics[width=\textwidth]{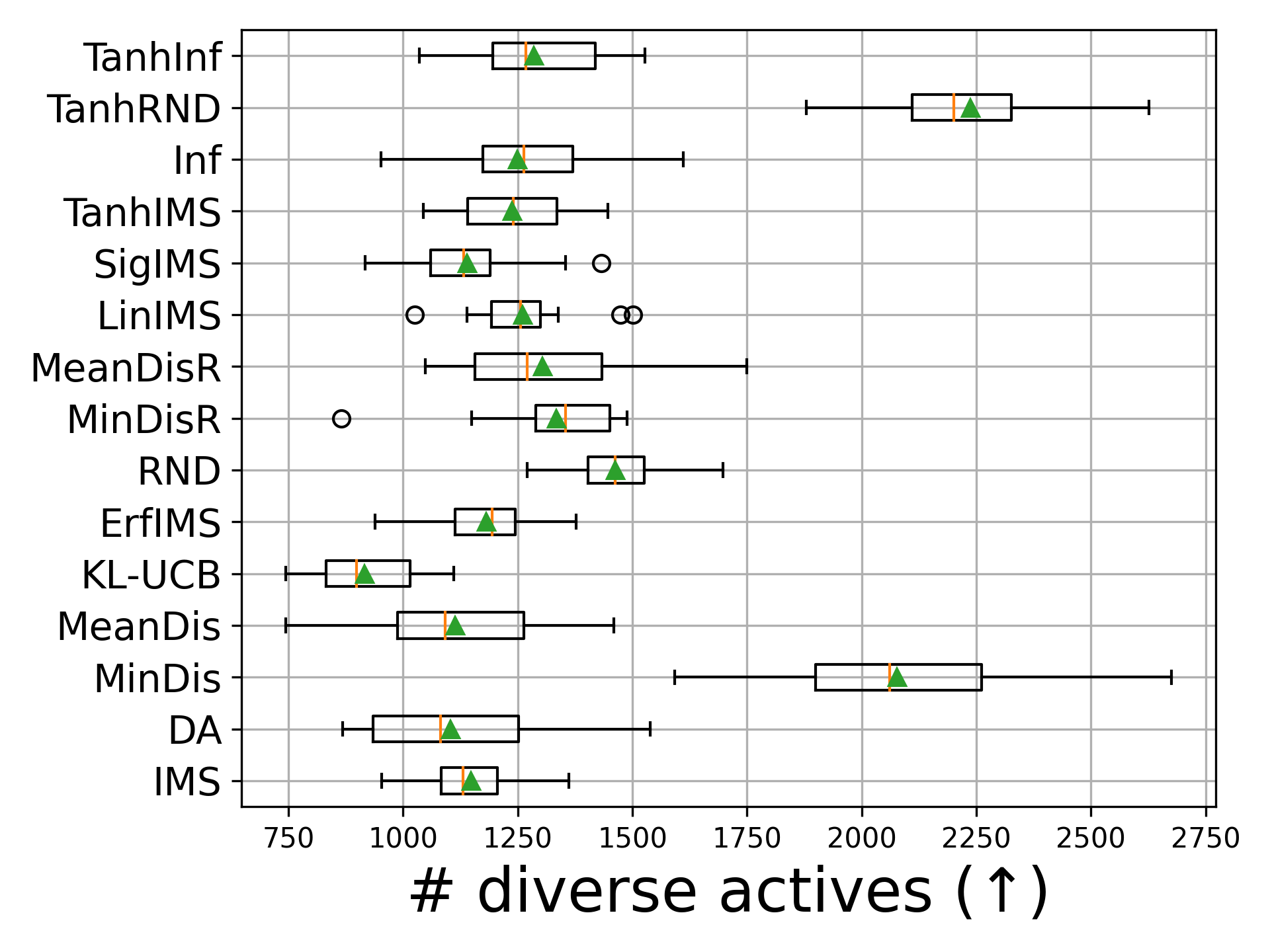}
         \caption{Diverse actives}
         \label{fig:gsk3b_circles}
     \end{subfigure}
        \caption{Evaluation on the GSK3$\beta$ oracle.}
        \label{fig:gsk3b}
\end{figure}
For the GSK3$\beta$ oracle, \cref{fig:gsk3b_reward} compares the average extrinsic reward (moving average using a window size of 101), over 20 independent runs, per generative step. We observe that the extrinsic rewards converge to comparable values across the diversity-aware reward functions. 
To evaluate the diversity, \cref{fig:gsk3b_chemical,fig:gsk3b_topo,fig:gsk3b_circles} display boxplots comparing the number of molecular scaffolds, topological scaffolds and diverse actives, respectively, over 20 independent runs with a budget of 2000 generative steps. Only active molecules with an extrinsic reward of at least $h=0.5$ are displayed, where orange lines and green triangles display the median and mean, respectively. Inf and TanhInf generate substantially more molecular and topological scaffolds; whereas TanhRND is also among the top methods. Concerning molecular scaffolds, we observe that the 7 most promising methods have a lower variability,i.e., interquartile range, among the 20 reruns, while the other methods often show a higher variability. Hence, the most promising methods can consistently generate a large number of molecular scaffolds.  In terms of topological scaffolds, TanhRND exhibits both low variability and among the largest number of topological scaffolds, together with TanhInf and Inf. 

MinDis and TanhRND generate more diverse actives than the other methods we have investigated, but MinDis shows a higher variability among the reruns. Overall, we observe a higher variability for the number of topological scaffolds and diverse actives among the reruns. This can partially be explained by the fact that the scaffolds-based methods only consider molecular scaffolds, but more molecular scaffolds do not necessarily lead to more topological scaffolds and diverse actives. However, as we observe here, if utilized appropriately, it can still enhance the overall diversity.

In general, applying RND alone does not yield among the highest diversities, but the combination of TanhIMS and RND can improve the diversity of TanhIMS, especially for topological and diverse actives. Hence, we observe that proper structure-based scaling of the extrinsic reward can help to improve diversity, while adding a prediction-based intrinsic reward, i.e., RND, can further enhance and stabilize diversity. Combining Inf, a structure-based intrinsic reward, with TanhIMS does not seem to enhance the diversity further, but it also does not reduce the diversity. We also observe enhanced performance in scaffold diversity for the non-binary penalty functions alone. Still, this structural information does not improve the similarity-based diversity, i.e., diverse actives.
 
\paragraph{JNK3}
\begin{figure}[h]
     \centering
     \begin{subfigure}[b]{0.23\textwidth}
         \centering
         \includegraphics[width=\textwidth]{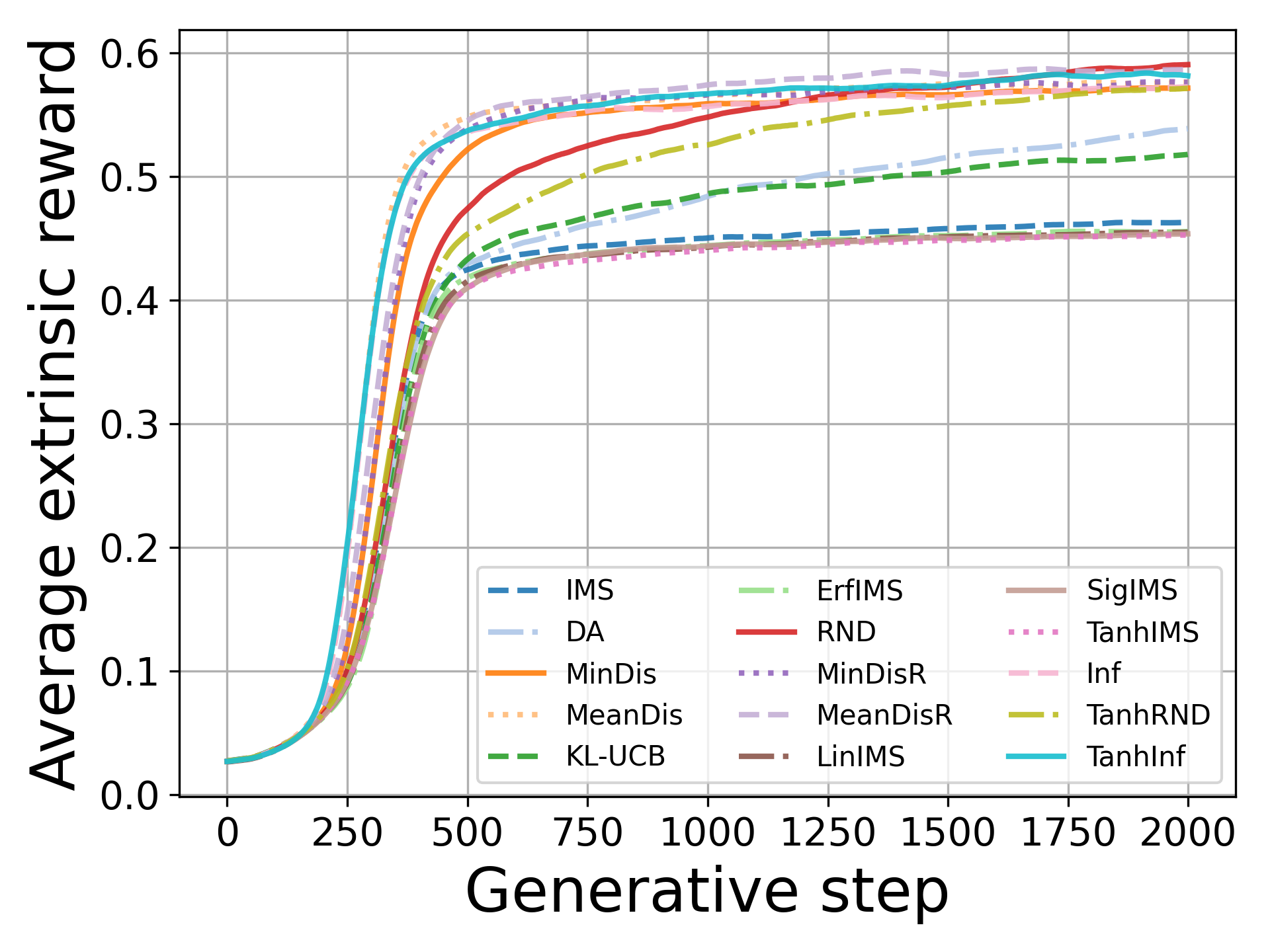}
         \caption{Extrinsic reward}
         \label{fig:jnk3_reward}
     \end{subfigure}
     \hfill
     \begin{subfigure}[b]{0.23\textwidth}
         \centering
         \includegraphics[width=\textwidth]{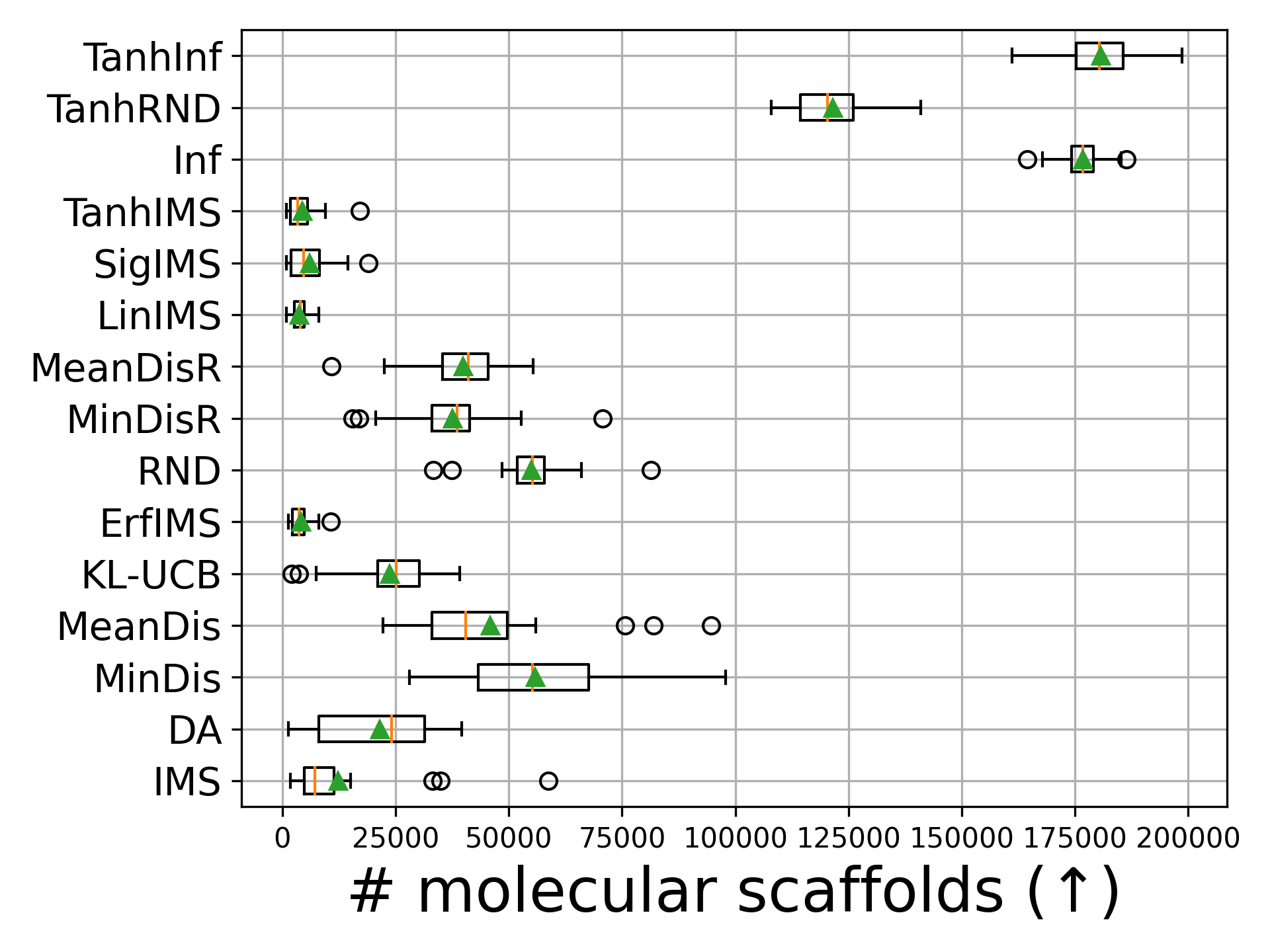}
         \caption{Molecular Scaffold}
         \label{fig:jnk3_chemical}
     \end{subfigure}
     \begin{subfigure}[b]{0.23\textwidth}
         \centering
         \includegraphics[width=\textwidth]{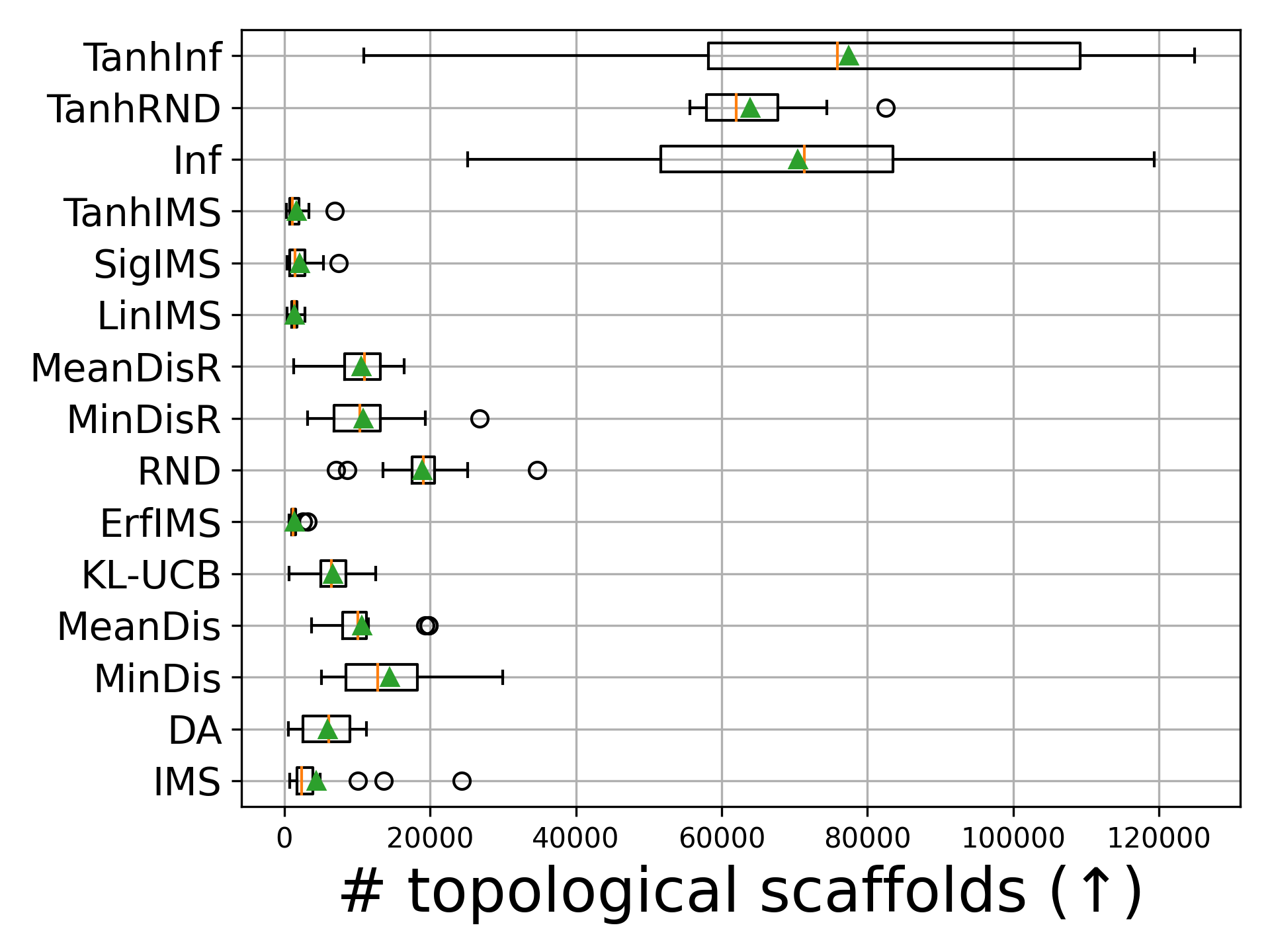}
         \caption{Topological scaffold}
         \label{fig:jnk3_topo}
     \end{subfigure}
     \hfill
     \begin{subfigure}[b]{0.23\textwidth}
         \centering
         \includegraphics[width=\textwidth]{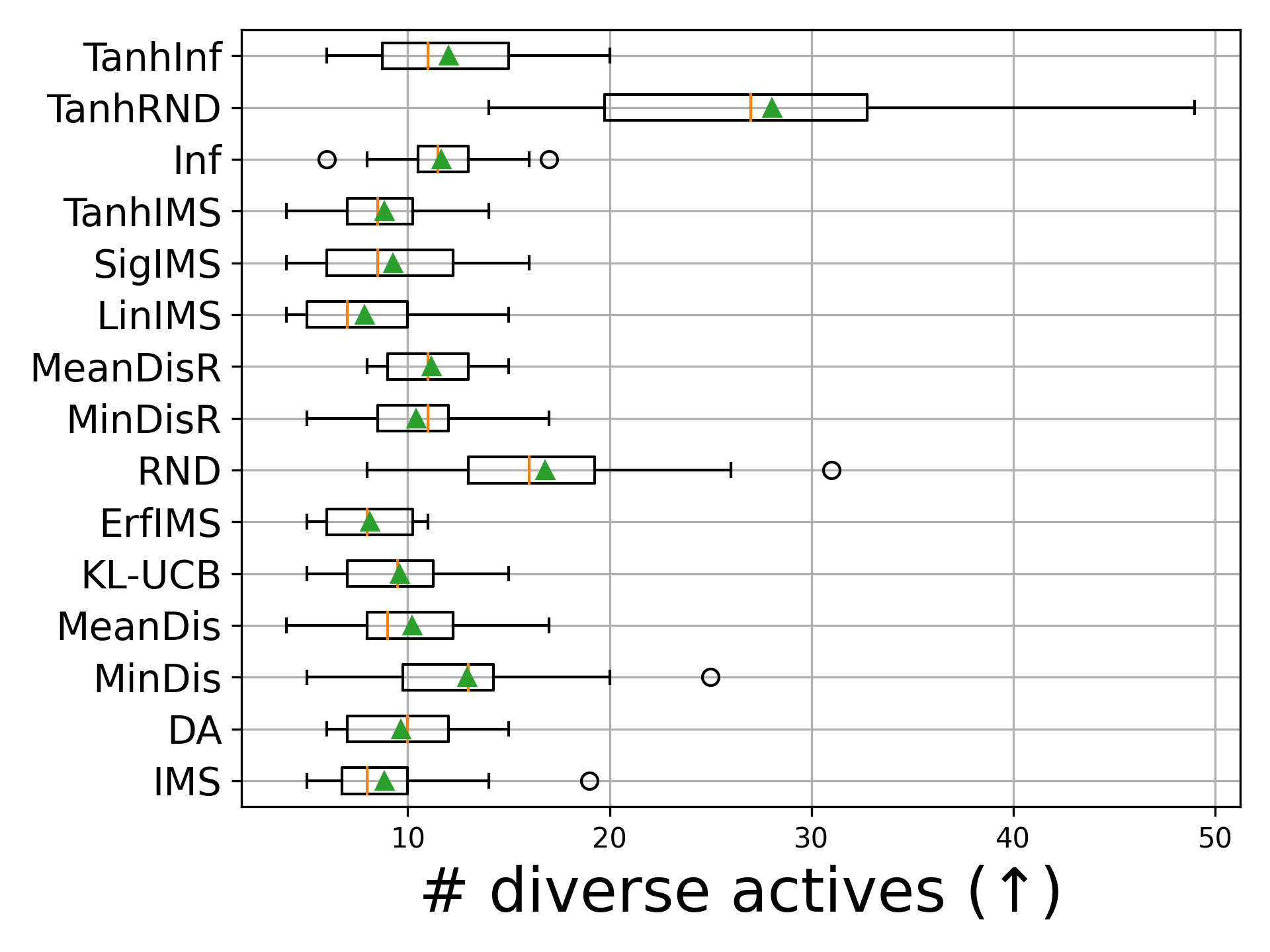}
         \caption{Diverse actives}
         \label{fig:jnk3_circles}
     \end{subfigure}
        \caption{Evaluation on the JNK3 oracle.}
        \label{fig:jnk3}
\end{figure} 
\Cref{fig:jnk3_reward} displays the average extrinsic reward (moving average using a window size of 101), over 20 independent runs, per generative step for the JNK3 oracle. We observe a substantially lower extrinsic reward when not using any intrinsic motivation.  The extrinsic rewards are significantly lower compared to the GSK3$\beta$ experiments, highlighting that this is a more difficult optimization problem where the agent is more prone to get stuck in a local optimum if not sufficient exploration is carried out. Note that the average extrinsic rewards of the penalty-based methods do not reach the threshold of $h=0.5$, where extrinsic rewards are penalized. However, by combining the penalty with an intrinsic reward, it is possible to escape low-rewarding local optima.  

\begin{figure}[h]
     \centering
     \begin{subfigure}[b]{0.23\textwidth}
         \centering
         \includegraphics[width=\textwidth]{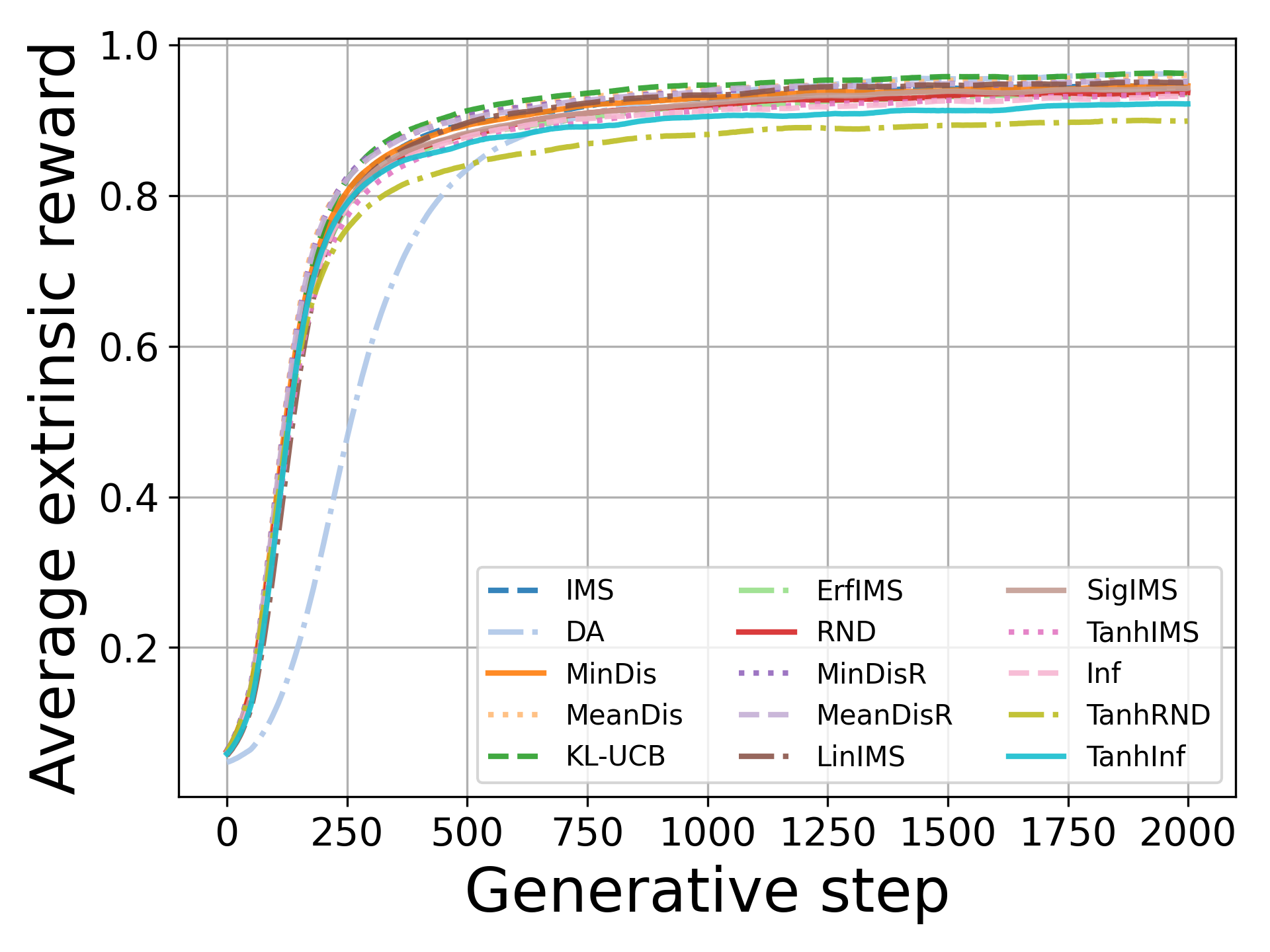}
         \caption{Extrinsic reward}
         \label{fig:drd2_reward}
     \end{subfigure}
     \hfill
     \begin{subfigure}[b]{0.23\textwidth}
         \centering
         \includegraphics[width=\textwidth]{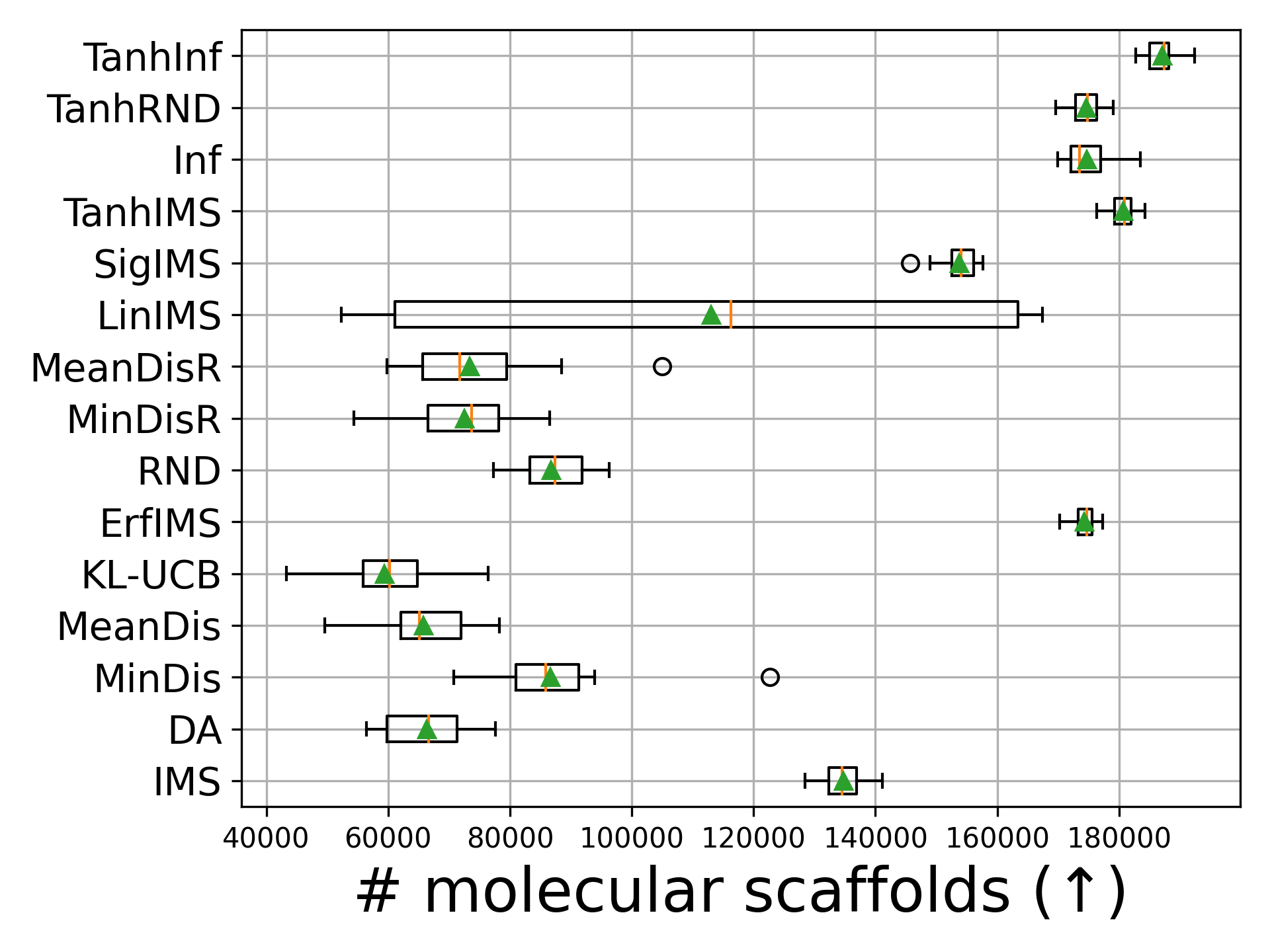}
         \caption{Molecular Scaffold}
         \label{fig:drd2_chemical}
     \end{subfigure}
     \begin{subfigure}[b]{0.23\textwidth}
         \centering
         \includegraphics[width=\textwidth]{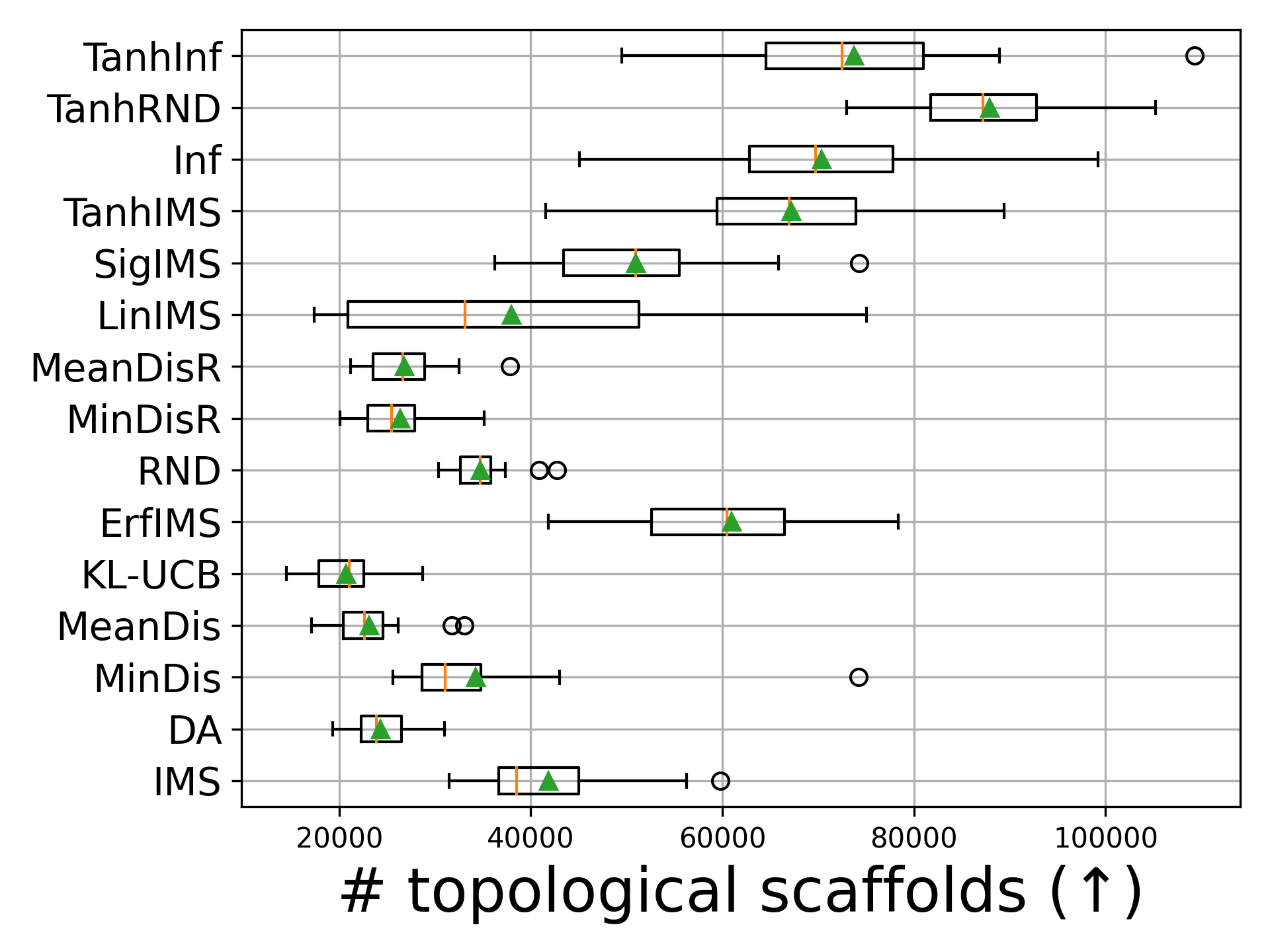}
         \caption{Topological scaffold}
         \label{fig:drd2_topo}
     \end{subfigure}
     \hfill
     \begin{subfigure}[b]{0.23\textwidth}
         \centering
         \includegraphics[width=\textwidth]{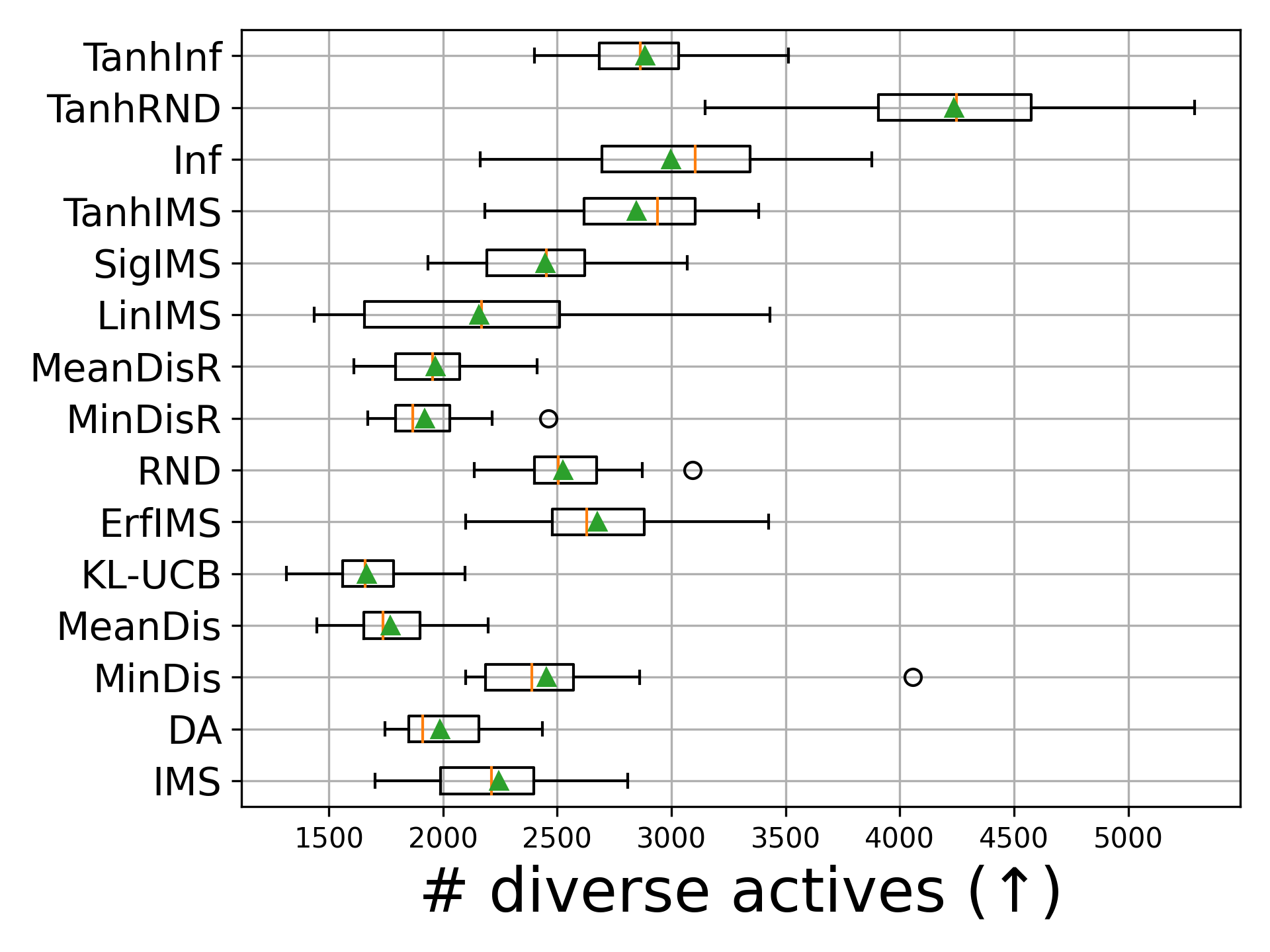}
         \caption{Diverse actives}
         \label{fig:drd2_circles}
     \end{subfigure}
        \caption{Evaluation on the DRD2 oracle.}
        \label{fig:drd2}
\end{figure}

We will now also see how the diversity of the generation process has been affected by the different diversity-aware reward functions. \Cref{fig:jnk3_chemical,fig:jnk3_topo,fig:jnk3_circles} show boxplots displaying the total number of molecular scaffolds, topological scaffolds, and diverse actives, respectively, after 2000 generative steps of 20 reruns. Only active molecules are considered, where orange lines and green triangles display the median and mean, respectively. Generally, substantially fewer molecular scaffolds, topological scaffolds, and diverse actives are found compared to the experiments on the GSK3$\beta$ extrinsic reward function (see \cref{fig:gsk3b_chemical,fig:gsk3b_topo,fig:gsk3b_circles}). This is because the JNK3-based extrinsic reward is more difficult to optimize (see \cref{fig:jnk3_reward,fig:gsk3b_reward,fig:drd2_reward}) and, therefore, the number of actives is lower. 
In terms of molecular and topological scaffolds (see \cref{fig:jnk3_chemical,fig:jnk3_topo}), it is evident that the methods that only penalize the extrinsic reward generate less diverse molecules. We observe that TanhInf, TanhRND and Inf generate substantially more molecular and topological scaffolds while their boxes (in the boxplots) do not intersect with the other methods. However, for TanhInf and Inf, we observe a higher variability in terms of topological scaffolds, while TanhRND demonstrates a more consistent number of topological scaffolds among the 20 reruns.

The differences between the methods are not as apparent in terms of diverse actives where most methods generate around 20 diverse actives. Only TanhRND shows the capability to generate more than 30 diverse actives. We notice that TanhRND has a higher variability in terms of diverse actives, but its box does not intersect with the other methods. The higher variability of TanhRND is possibly because it is difficult to improve the number of diverse actives and, therefore, there is a natural variability in how much it can be improved in each rerun.

Overall, TanhRND shows a high diversity, with only TanhInf and Inf showing significantly more molecular scaffolds. On the other hand, for RND and TanhIMS on their own, we observe no substantial improvement in diversity compared to the other methods. This is likely because JNK3 is more difficult to optimize (see Figure \ref{fig:jnk3_reward}) due to having fewer optima compared to GSK3$\beta$. Hence, we observe that the combination of RND and TanhIMS considerably improves the overall diversity for a more sparse optimization task.

\paragraph{DRD2}
\cref{fig:drd2_reward} shows the average extrinsic reward (moving average using a window size of 101), over 20 independent runs, per generative step for the DRD2 oracle. We observe that the extrinsic rewards converge to comparable values across the diversity-aware reward functions. All reward functions exhibits larger average extrinsic rewards than both the JNK3 and GSK3$\beta$ oracle (see \cref{fig:gsk3b_reward,fig:jnk3_reward}). Hence, DRD2 seems to be easier to optimize and, therefore, we observe no substantial difference between the diversity-aware reward functions in terms extrinsic reward.    
To evaluate the diversity on the DRD2 oracle, \cref{fig:drd2_chemical,fig:drd2_topo,fig:drd2_circles} displays boxplots comparing the total number of molecular scaffolds, topological scaffolds, and diverse actives after 2000 generative steps across 20 independent runs. Only active molecules with an extrinsic reward of at least $h=0.5$ are displayed. In terms of molecular scaffolds, we observe that the best methods show a lower variability, consistently generating substantially more molecular scaffolds. Concerning topological scaffolds and diverse actives, the best methods show a higher or similar variability compared to the other methods. As mentioned above, this is likely because a higher number of molecular scaffolds does not directly lead to more topological scaffolds and diverse actives. However, as we observe here, it can still enhance the overall diversity if used appropriately.  

Overall, TanhRND consistently ranks among the top methods and frequently generates more topological scaffolds and diverse actives than the other methods. It shows capabilities to generate more than 4000 diverse actives, which is substantially more than the other diversity-aware reward functions. 


\section{Conclusion}
Our comprehensive study proposes and evaluates several novel intrinsic rewards and reward penalties to enhance the diversity of \textit{de novo} drug design using reinforcement learning (RL). Our approach balances exploration and exploitation to promote a more diverse generation of molecules. 
Our results consistently show that methods incorporating both intrinsic reward and reward penalty generate significantly more diverse actives, molecular scaffolds, and topological scaffolds. In particular, using structure-based scaling of the extrinsic reward and a prediction-based method to encourage exploration in the agent's state space, we observe improved diversity in terms of both structure and similarity. Yet, a single type of information does not fully enhance diversity. Our work opens several future directions for studying how to incorporate domain and agent information into the reward signal efficiently.


\section*{Acknowledgments}
This work was partially supported by the Wallenberg AI, Autonomous Systems and Software Program (WASP) funded by the Knut and Alice Wallenberg Foundation. The experimental evaluation was enabled by resources provided by the National Academic Infrastructure for Supercomputing in Sweden (NAISS), partially funded by the Swedish Research Council through grant agreement no. 2022-06725.


\bibliographystyle{named}
\bibliography{main}

\begin{thebibliography}{}

\bibitem[\protect\citeauthoryear{Atance \bgroup \em et al.\egroup }{2022}]{atance2022novo}
Sara~Romeo Atance, Juan~Viguera Diez, Ola Engkvist, Simon Olsson, and Roc{\'\i}o Mercado.
\newblock De novo drug design using reinforcement learning with graph-based deep generative models.
\newblock {\em Journal of chemical information and modeling}, 62(20):4863--4872, 2022.

\bibitem[\protect\citeauthoryear{Badia \bgroup \em et al.\egroup }{2020}]{badia2020uplearningdirectedexploration}
Adrià~Puigdomènech Badia, Pablo Sprechmann, Alex Vitvitskyi, Daniel Guo, Bilal Piot, Steven Kapturowski, Olivier Tieleman, Martín Arjovsky, Alexander Pritzel, Andew Bolt, and Charles Blundell.
\newblock Never give up: Learning directed exploration strategies, 2020.

\bibitem[\protect\citeauthoryear{Bellemare \bgroup \em et al.\egroup }{2016}]{bellemare2016unifying}
Marc~G. Bellemare, Sriram Srinivasan, Georg Ostrovski, Tom Schaul, David Saxton, and R\'{e}mi Munos.
\newblock Unifying count-based exploration and intrinsic motivation.
\newblock In {\em Proceedings of the 30th International Conference on Neural Information Processing Systems}, NIPS'16, page 1479–1487, Red Hook, NY, USA, 2016. Curran Associates Inc.

\bibitem[\protect\citeauthoryear{Bemis and Murcko}{1996}]{bemis1996properties}
Guy~W Bemis and Mark~A Murcko.
\newblock The properties of known drugs. 1. molecular frameworks.
\newblock {\em Journal of medicinal chemistry}, 39(15):2887--2893, 1996.

\bibitem[\protect\citeauthoryear{Blaschke \bgroup \em et al.\egroup }{2020a}]{blaschke2020reinvent}
Thomas Blaschke, Josep Ar{\'u}s-Pous, Hongming Chen, Christian Margreitter, Christian Tyrchan, Ola Engkvist, Kostas Papadopoulos, and Atanas Patronov.
\newblock Reinvent 2.0: an ai tool for de novo drug design.
\newblock {\em Journal of chemical information and modeling}, 60(12):5918--5922, 2020.

\bibitem[\protect\citeauthoryear{Blaschke \bgroup \em et al.\egroup }{2020b}]{blaschke2020memory}
Thomas Blaschke, Ola Engkvist, J{\"u}rgen Bajorath, and Hongming Chen.
\newblock Memory-assisted reinforcement learning for diverse molecular de novo design.
\newblock {\em Journal of cheminformatics}, 12(1):68, 2020.

\bibitem[\protect\citeauthoryear{Burda \bgroup \em et al.\egroup }{2018}]{burda2018exploration}
Yuri Burda, Harrison Edwards, Amos Storkey, and Oleg Klimov.
\newblock Exploration by random network distillation.
\newblock {\em arXiv preprint arXiv:1810.12894}, 2018.

\bibitem[\protect\citeauthoryear{DiMasi \bgroup \em et al.\egroup }{2016}]{dimasi2016innovation}
Joseph~A DiMasi, Henry~G Grabowski, and Ronald~W Hansen.
\newblock Innovation in the pharmaceutical industry: new estimates of r\&d costs.
\newblock {\em Journal of health economics}, 47:20--33, 2016.

\bibitem[\protect\citeauthoryear{Gao \bgroup \em et al.\egroup }{2022}]{gao2022sample}
Wenhao Gao, Tianfan Fu, Jimeng Sun, and Connor~W. Coley.
\newblock Sample efficiency matters: {A} benchmark for practical molecular optimization.
\newblock In Sanmi Koyejo, S.~Mohamed, A.~Agarwal, Danielle Belgrave, K.~Cho, and A.~Oh, editors, {\em Advances in Neural Information Processing Systems 35: Annual Conference on Neural Information Processing Systems 2022, NeurIPS 2022, New Orleans, LA, USA, November 28 - December 9, 2022}, 2022.

\bibitem[\protect\citeauthoryear{Garivier and Cappé}{2011}]{pmlr-v19-garivier11a}
Aurélien Garivier and Olivier Cappé.
\newblock The kl-ucb algorithm for bounded stochastic bandits and beyond.
\newblock In Sham~M. Kakade and Ulrike von Luxburg, editors, {\em Proceedings of the 24th Annual Conference on Learning Theory}, volume~19 of {\em Proceedings of Machine Learning Research}, pages 359--376, Budapest, Hungary, 09--11 Jun 2011. PMLR.

\bibitem[\protect\citeauthoryear{Gaulton \bgroup \em et al.\egroup }{2017}]{gaulton2017chembl}
Anna Gaulton, Anne Hersey, Micha{\l} Nowotka, A~Patricia Bento, Jon Chambers, David Mendez, Prudence Mutowo, Francis Atkinson, Louisa~J Bellis, Elena Cibri{\'a}n-Uhalte, et~al.
\newblock The chembl database in 2017.
\newblock {\em Nucleic acids research}, 45(D1):D945--D954, 2017.

\bibitem[\protect\citeauthoryear{Gummesson~Svensson \bgroup \em et al.\egroup }{2024}]{gummesson2024utilizing}
Hampus Gummesson~Svensson, Christian Tyrchan, Ola Engkvist, and Morteza Haghir~Chehreghani.
\newblock Utilizing reinforcement learning for de novo drug design.
\newblock {\em Machine Learning}, 113(7):4811--4843, 2024.

\bibitem[\protect\citeauthoryear{Guo and Schwaller}{2024}]{guo2024augmented}
Jeff Guo and Philippe Schwaller.
\newblock Augmented memory: Sample-efficient generative molecular design with reinforcement learning.
\newblock {\em Jacs Au}, 2024.

\bibitem[\protect\citeauthoryear{Hochreiter}{1997}]{hochreiter1997long}
S~Hochreiter.
\newblock Long short-term memory.
\newblock {\em Neural Computation MIT-Press}, 1997.

\bibitem[\protect\citeauthoryear{Hughes \bgroup \em et al.\egroup }{2011}]{hughes2011principles}
James~P Hughes, Stephen Rees, S~Barrett Kalindjian, and Karen~L Philpott.
\newblock Principles of early drug discovery.
\newblock {\em British journal of pharmacology}, 162(6):1239--1249, 2011.

\bibitem[\protect\citeauthoryear{Jaccard}{1912}]{jaccard1912distribution}
Paul Jaccard.
\newblock The distribution of the flora in the alpine zone. 1.
\newblock {\em New phytologist}, 11(2):37--50, 1912.

\bibitem[\protect\citeauthoryear{Jasial \bgroup \em et al.\egroup }{2016}]{10.12688/f1000research.8357.2}
S~Jasial, Y~Hu, M~Vogt, and J~Bajorath.
\newblock Activity-relevant similarity values for fingerprints and implications for similarity searching.
\newblock {\em F1000Research}, 5(591), 2016.

\bibitem[\protect\citeauthoryear{Kingma and Ba}{2017}]{kingma2017adammethodstochasticoptimization}
Diederik~P. Kingma and Jimmy Ba.
\newblock Adam: A method for stochastic optimization, 2017.

\bibitem[\protect\citeauthoryear{Landrum}{2006}]{landrum2006rdkit}
Greg Landrum.
\newblock Rdkit: Open-source cheminformatics.
\newblock \url{http://www.rdkit.org}, 2006.

\bibitem[\protect\citeauthoryear{Laud}{2004}]{laud2004theory}
Adam~Daniel Laud.
\newblock {\em Theory and application of reward shaping in reinforcement learning}.
\newblock PhD thesis, University of Illinois at Urbana-Champaign, 2004.

\bibitem[\protect\citeauthoryear{Liu \bgroup \em et al.\egroup }{2021}]{liu2021drugex}
Xuhan Liu, Kai Ye, Herman~WT van Vlijmen, Michael~TM Emmerich, Adriaan~P IJzerman, and Gerard~JP van Westen.
\newblock Drugex v2: de novo design of drug molecules by pareto-based multi-objective reinforcement learning in polypharmacology.
\newblock {\em Journal of cheminformatics}, 13(1):85, 2021.

\bibitem[\protect\citeauthoryear{Loeffler \bgroup \em et al.\egroup }{2024}]{loeffler2024reinvent}
Hannes~H Loeffler, Jiazhen He, Alessandro Tibo, Jon~Paul Janet, Alexey Voronov, Lewis~H Mervin, and Ola Engkvist.
\newblock Reinvent 4: Modern ai--driven generative molecule design.
\newblock {\em Journal of Cheminformatics}, 16(1):20, 2024.

\bibitem[\protect\citeauthoryear{Olivecrona \bgroup \em et al.\egroup }{2017}]{olivecrona2017molecular}
Marcus Olivecrona, Thomas Blaschke, Ola Engkvist, and Hongming Chen.
\newblock Molecular de-novo design through deep reinforcement learning.
\newblock {\em Journal of cheminformatics}, 9:1--14, 2017.

\bibitem[\protect\citeauthoryear{Park \bgroup \em et al.\egroup }{2024}]{park2024molairmolecularreinforcementlearning}
Jinyeong Park, Jaegyoon Ahn, Jonghwan Choi, and Jibum Kim.
\newblock Mol-air: Molecular reinforcement learning with adaptive intrinsic rewards for goal-directed molecular generation, 2024.

\bibitem[\protect\citeauthoryear{Pitt \bgroup \em et al.\egroup }{2025}]{pitt2025real}
Will~R. Pitt, Jonathan Bentley, Christophe Boldron, Lionel Colliandre, Carmen Esposito, Elizabeth~H. Frush, Jola Kopec, St{\'e}phanie Labouille, Jerome Meneyrol, David~A. Pardoe, Ferruccio Palazzesi, Alfonso Pozzan, Jacob~M. Remington, Ren{\'e} Rex, Michelle Southey, Sachin Vishwakarma, and Paul Walker.
\newblock Real-world applications and experiences of ai/ml deployment for drug discovery.
\newblock {\em Journal of Medicinal Chemistry}, 2025.
\newblock PMID: 39772505.

\bibitem[\protect\citeauthoryear{Popova \bgroup \em et al.\egroup }{2018}]{popova2018deep}
Mariya Popova, Olexandr Isayev, and Alexander Tropsha.
\newblock Deep reinforcement learning for de novo drug design.
\newblock {\em Science advances}, 4(7):eaap7885, 2018.

\bibitem[\protect\citeauthoryear{Quancard \bgroup \em et al.\egroup }{2023}]{quancard2023european}
Jean Quancard, Anna Vulpetti, Anders Bach, Brian Cox, St{\'e}phanie~M Gu{\'e}ret, Ingo~V Hartung, Hannes~F Koolman, Stefan Laufer, Josef Messinger, Gianluca Sbardella, et~al.
\newblock The european federation for medicinal chemistry and chemical biology (efmc) best practice initiative: Hit generation.
\newblock {\em ChemMedChem}, 18(9):e202300002, 2023.

\bibitem[\protect\citeauthoryear{Renz \bgroup \em et al.\egroup }{2019}]{renz2019failure}
Philipp Renz, Dries Van~Rompaey, J{\"o}rg~Kurt Wegner, Sepp Hochreiter, and G{\"u}nter Klambauer.
\newblock On failure modes in molecule generation and optimization.
\newblock {\em Drug Discovery Today: Technologies}, 32:55--63, 2019.

\bibitem[\protect\citeauthoryear{Renz \bgroup \em et al.\egroup }{2024}]{renz2024diverse}
Philipp Renz, Sohvi Luukkonen, and Gu{\"u}nter Klambauer.
\newblock Diverse hits in de novo molecule design: Diversity-based comparison of goal-directed generators.
\newblock {\em Journal of Chemical Information and Modeling}, 64(15):5756--5761, 2024.

\bibitem[\protect\citeauthoryear{Reymond}{2015}]{reymond2015chemical}
Jean-Louis Reymond.
\newblock The chemical space project.
\newblock {\em Accounts of chemical research}, 48(3):722--730, 2015.

\bibitem[\protect\citeauthoryear{Rogers and Hahn}{2010}]{rogers2010extended}
David Rogers and Mathew Hahn.
\newblock Extended-connectivity fingerprints.
\newblock {\em Journal of chemical information and modeling}, 50(5):742--754, 2010.

\bibitem[\protect\citeauthoryear{Rumelhart \bgroup \em et al.\egroup }{1985}]{rumelhart1985learning}
David~E Rumelhart, Geoffrey~E Hinton, and Ronald~J Williams.
\newblock Learning internal representations by error propagation.
\newblock Technical report, California Univ San Diego La Jolla Inst for Cognitive Science, 1985.

\bibitem[\protect\citeauthoryear{Segler \bgroup \em et al.\egroup }{2018}]{segler2018generating}
Marwin~HS Segler, Thierry Kogej, Christian Tyrchan, and Mark~P Waller.
\newblock Generating focused molecule libraries for drug discovery with recurrent neural networks.
\newblock {\em ACS central science}, 4(1):120--131, 2018.

\bibitem[\protect\citeauthoryear{Sun \bgroup \em et al.\egroup }{2017}]{sun2017excape}
Jiangming Sun, Nina Jeliazkova, Vladimir Chupakhin, Jose-Felipe Golib-Dzib, Ola Engkvist, Lars Carlsson, J{\"o}rg Wegner, Hugo Ceulemans, Ivan Georgiev, Vedrin Jeliazkov, et~al.
\newblock Excape-db: an integrated large scale dataset facilitating big data analysis in chemogenomics.
\newblock {\em Journal of cheminformatics}, 9:1--9, 2017.

\bibitem[\protect\citeauthoryear{Thomas \bgroup \em et al.\egroup }{2022a}]{thomas2022re}
Morgan Thomas, Noel~M. O'Boyle, Andreas Bender, and Chris~De Graaf.
\newblock Re-evaluating sample efficiency in de novo molecule generation, 2022.

\bibitem[\protect\citeauthoryear{Thomas \bgroup \em et al.\egroup }{2022b}]{thomas2022augmented}
Morgan Thomas, Noel~M O’Boyle, Andreas Bender, and Chris De~Graaf.
\newblock Augmented hill-climb increases reinforcement learning efficiency for language-based de novo molecule generation.
\newblock {\em Journal of cheminformatics}, 14(1):68, 2022.

\bibitem[\protect\citeauthoryear{Velez-Arce \bgroup \em et al.\egroup }{2024}]{Velez-Arce2024tdc}
Alejandro Velez-Arce, Kexin Huang, Michelle Li, Xiang Lin, Wenhao Gao, Tianfan Fu, Manolis Kellis, Bradley~L. Pentelute, and Marinka Zitnik.
\newblock Tdc-2: Multimodal foundation for therapeutic science.
\newblock {\em bioRxiv}, 2024.

\bibitem[\protect\citeauthoryear{Wang and Zhu}{2024}]{wang2024exselfrl}
Jing Wang and Fei Zhu.
\newblock Exselfrl: An exploration-inspired self-supervised reinforcement learning approach to molecular generation.
\newblock {\em Expert Systems with Applications}, page 125410, 2024.

\bibitem[\protect\citeauthoryear{Weininger}{1988}]{weininger1988smiles}
David Weininger.
\newblock Smiles, a chemical language and information system. 1. introduction to methodology and encoding rules.
\newblock {\em Journal of chemical information and computer sciences}, 28(1):31--36, 1988.

\bibitem[\protect\citeauthoryear{Wouters \bgroup \em et al.\egroup }{2020}]{wouters2020estimated}
Olivier~J Wouters, Martin McKee, and Jeroen Luyten.
\newblock Estimated research and development investment needed to bring a new medicine to market, 2009-2018.
\newblock {\em Jama}, 323(9):844--853, 2020.

\bibitem[\protect\citeauthoryear{Xie \bgroup \em et al.\egroup }{2023}]{xie2023much}
Yutong Xie, Ziqiao Xu, Jiaqi Ma, and Qiaozhu Mei.
\newblock How much space has been explored? measuring the chemical space covered by databases and machine-generated molecules.
\newblock In {\em The Eleventh International Conference on Learning Representations, {ICLR} 2023, Kigali, Rwanda, May 1-5, 2023}. OpenReview.net, 2023.

\end{thebibliography}


\clearpage
\appendix
\section{Experimental 
Details}
\label{app:experimental_details}
The policy $\pi_{\theta}$ is a neural network with an embedding layer and a subsequent multi-layer long short-term memory (LSTM) \citep{hochreiter1997long} recurrent neural network (RNN). The policy's action probabilities are obtained by feeding the LSTM output through a fully connected layer and a subsequent softmax layer. Finetuning of the policy network is done on a single NVIDIA A40 GPU with 48GB RAM using PyTorch 2.4.1 and CUDA 12.4. At the end of each generative step, the parameters of the embedding, LSTM, and fully-connected layers are updated by performing one gradient step on the generated batch of molecules. To perform a gradient step update, we use Adam\citep{kingma2017adammethodstochasticoptimization} with a learning rate of $10^{-4}$ and keep other default parameters in Adam. Oracle functions, providing the extrinsic rewards, provided by PyTDC 0.4.17. Fingerprints are computed using RDKit 2023.9.6. 
Parameter $\sigma$ of the augmented likelihood is automatically adjusted as described in \cref{app:adaptive_sigma}, initialized to the value of $\sigma_{\text{init}}$. Hyperparameters utilized in the experiments are displayed in \cref{tab:parameters}. The source code is available as part of a framework for SMILES-based \textit{de novo} drug design\footnote{\url{https://github.com/MolecularAI/SMILES-RL}}.

\begin{table}[h!]
    \centering
    \caption{Parameters and corresponding values utilized in the experiments.}
    \label{tab:parameters}
    \begin{tabular}{c|c}
         Parameter & Value  \\ \hline
         Num. actions $|\mathcal{A}|$ & 34\\
         Extrinsic reward threshold $h$ & 0.5 \\
         KL-UCB parameter $c$ & 0 \\
         $c_{\text{tanh}}$ & 3 \\
         Distance threshold $D$ & 0.7 \\
         Bucket size $m$ & 25\\
         Batch size $|\mathcal{B}|$ & 128\\
         Num. generative steps $I$ & 2000\\
         Learning rate $\alpha$ & $10^{-4}$\\
         layer size & 512\\
         Num. recurrent layers & 3\\
         Embedding layer size & 256\\
         Optimizer & Adam\citep{kingma2017adammethodstochasticoptimization}\\
         $\sigma_{\text{init}}$ & 128 \\
         $m_{\sigma}$ & 50\\
         $w_{\sigma}$ & 10\\
          $D^{\min}_{\sigma}$ & 0.15\\
         $T_{\text{max}}$ & 256\\   
         Num. independent runs & 20\\ 
    \end{tabular}

\end{table}

\subsection{Automtic update of $\sigma$}
\label{app:adaptive_sigma}
The scalar parameter $\sigma$ of the augmented likelihood is automatically updated based on the difference between the agent likelihood and augmented likelihood. This was introduced in REINVENT 3.0\footnote{\url{https://github.com/MolecularAI/Reinvent/tree/v3.0}}, called margin guard. We follow the update procedure used in REINVENT 3.0, as described below.

For a generative step $i$, the difference between defined by
\begin{equation}
\begin{split}
   \delta_{\sigma} = \frac{1}{|\mathcal{K}_{i-1}|} \sum_{a_{1:T-2}\in \mathcal{K}_{i-1}} \left(\log \pi_{\theta_{\text{aug}}}(a_{1:T})\right.\\
   \left.- \sum_{t=1}^{T-2}\log \pi_{\theta}(a_t|s_t)\right),
   \end{split}
\end{equation}
where $\mathcal{K}_{i-1}$ is all molecules generated before generative step $i$. The $\sigma$ parameter is initialized to the value $\sigma_{\text{init}}$. After at least $w_{\sigma}$ generative steps, the parameter $\sigma$ is adjusted if $\delta_{\sigma} > m_{\sigma}$. Given desirable minimum score $D^{\min}_{\sigma}$, let 
\begin{equation}
    D_{\sigma} = \max\left(\frac{1}{|\mathcal{G}_{i-1}|} \sum_{a_{1:T-2}\in \mathcal{G}_{i-1}} r\left(a_{1:T-2}\right),D^{\min}_{\sigma}\right),
\end{equation}
where $\mathcal{G}_{i-1}$ is the set of previously generated molecules.
If $\sigma$ is updated, it is increased to
\begin{equation}
    \sigma = \max\left(\sigma, \frac{\delta_{\sigma}}{D_{\sigma}}\right) + m_{\sigma}.
\end{equation}
If $\sigma$ is adjusted, the weights $\theta$ of the policy $\pi_{\theta}$ are re-initialized to the pre-trained (prior) weights.


\subsection{Run time}
\begin{figure*}
     \centering
     \begin{subfigure}[b]{0.30\textwidth}
         \centering
         \includegraphics[width=\textwidth]{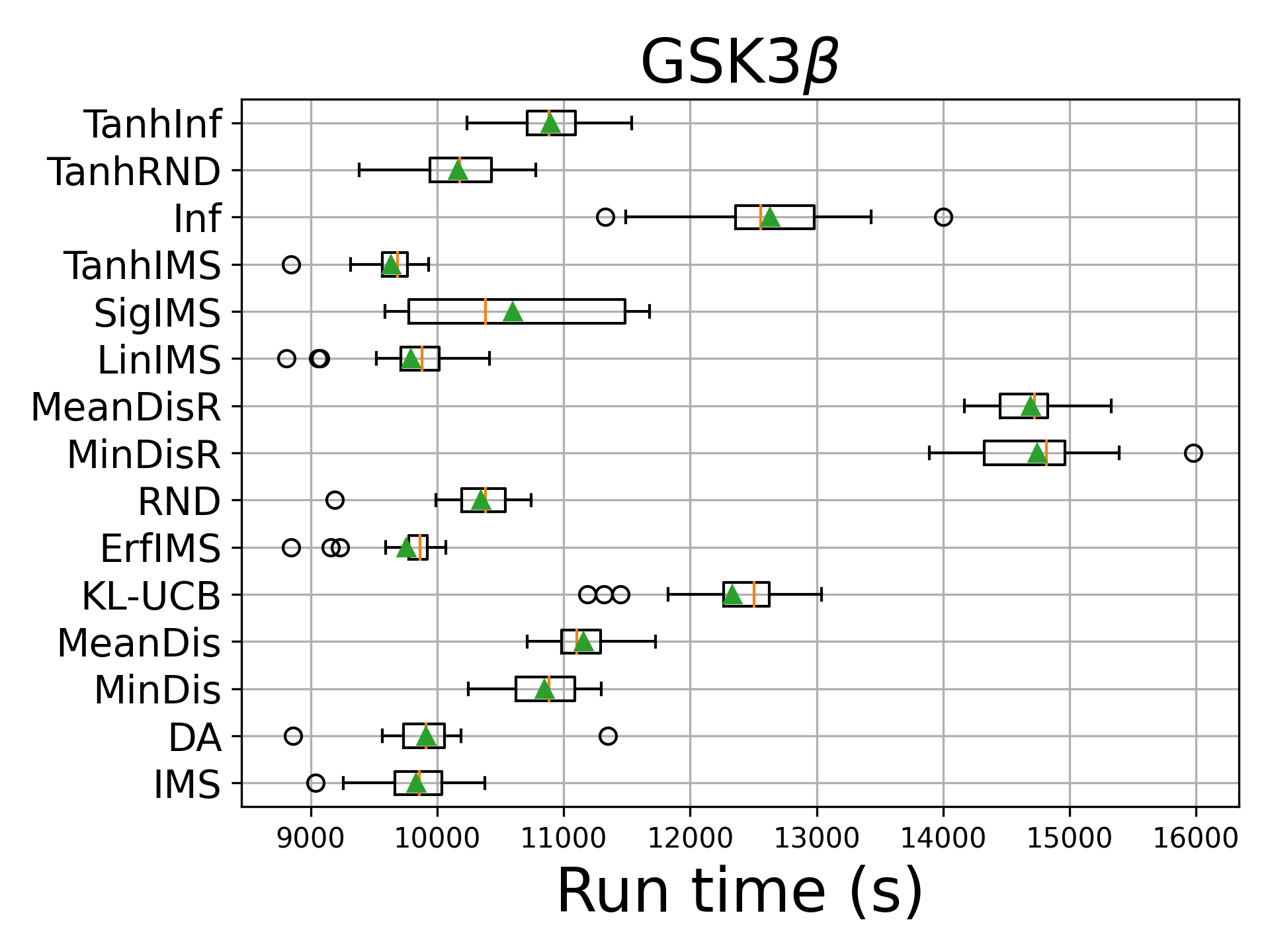}
         \caption{GSK3$\beta $}
         \label{fig:run_time_gsk3b}
     \end{subfigure}
     \hfill
     \begin{subfigure}[b]{0.30\textwidth}
         \centering
         \includegraphics[width=\textwidth]{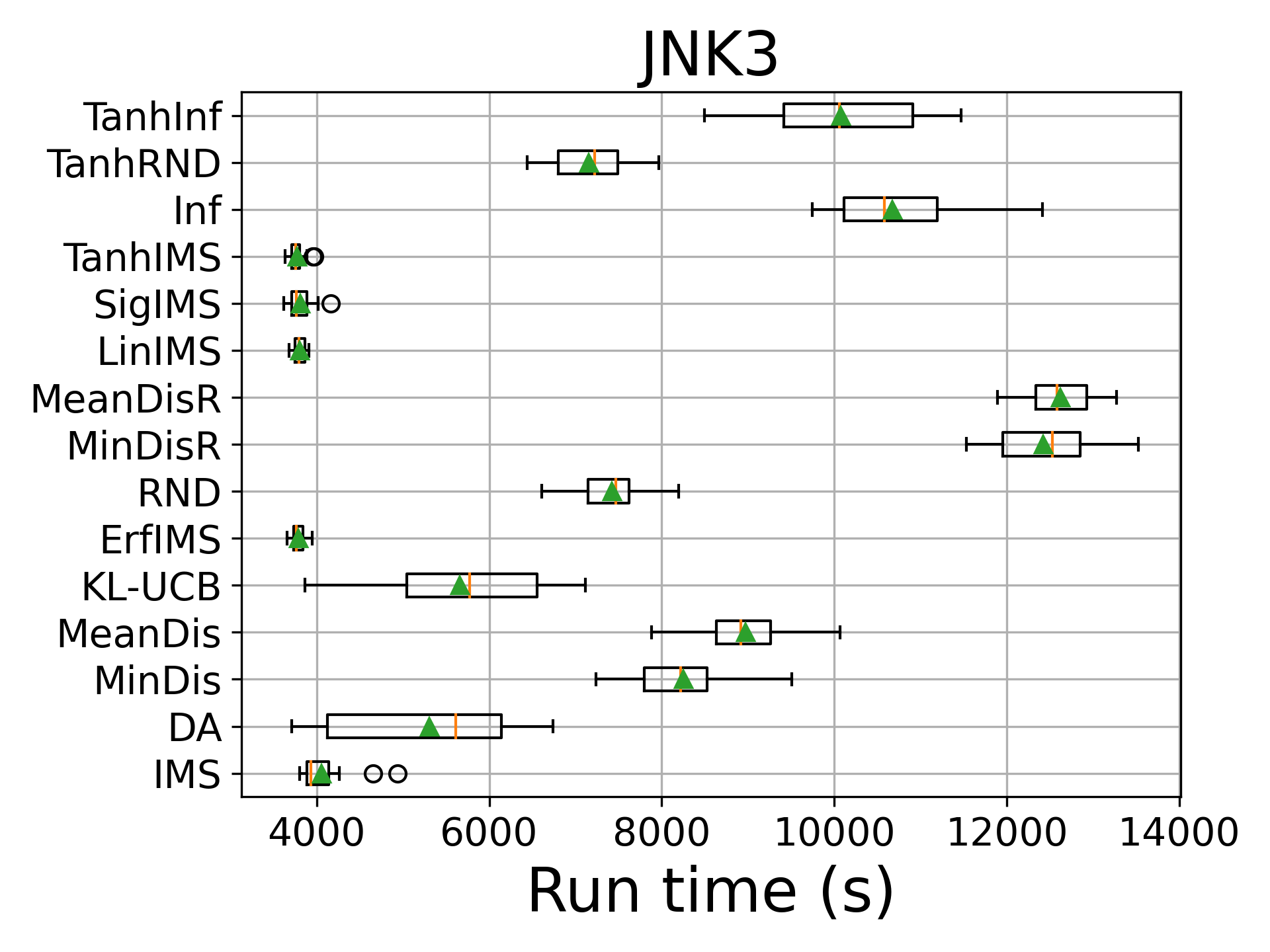}
         \caption{JNK3}
         \label{fig:runtime_jnk3}
     \end{subfigure}
     \hfill
     \begin{subfigure}[b]{0.30\textwidth}
         \centering
         \includegraphics[width=\textwidth]{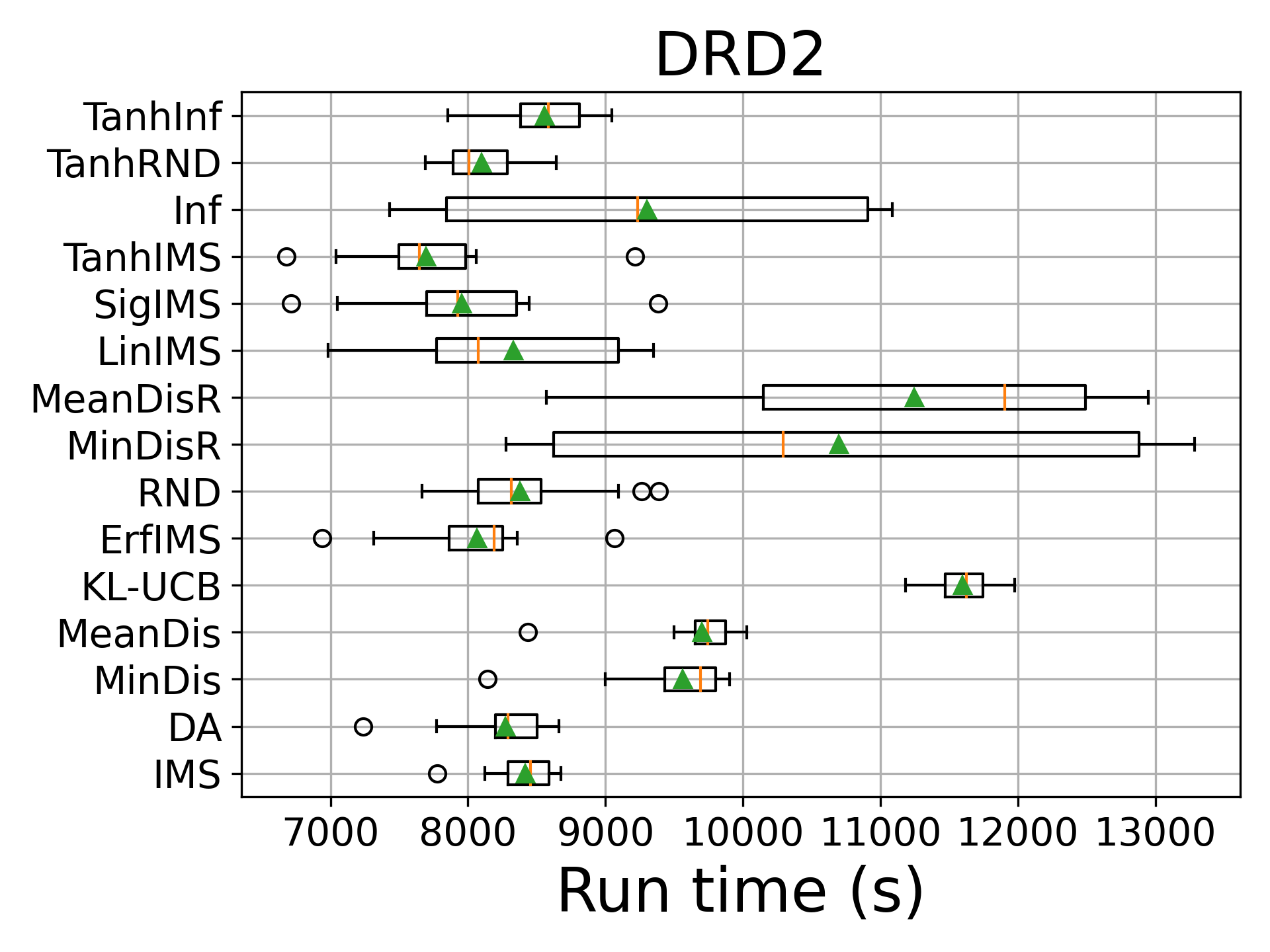}
         \caption{DRD2}
         \label{fig:runtime_drd2}
     \end{subfigure}
        \caption{Displays boxplots of the run time over 20 independent reruns for the different oracles.}
        \label{fig:run_time}
\end{figure*}

\Cref{fig:run_time} shows the runtimes over 20 independent reruns for the GSK3$\beta$, JNK3 and DRD2 oracle. As expected, the distance-based strategies generally display a longer run time, since they involve computing pair-wise distances against a (potentially large) set. The extrinsic penalty functions involve a conversion and lookup for scaffolds, which is usually scalable. Also, KL-UCB can require a long runtime since it involves solving an optimization problem at each iteration. Interestingly, the information-based approach, i.e., Inf, displays one of the longest runtimes. One possible explanation is that it consistently generates more scaffolds than the other scaffold-based strategies, leading to more scaffolds being in the memory and, consequently, a longer run time for the scaffold lookup. However, it needs to be further analyzed to establish the real cause.

\section{Diversity per Generative Step}
In this section, we display the Cumulative number of molecular scaffolds, Topological Scaffolds and Diverse Hits per generative step $i$. We display the mean and sample standard deviation over 20 independent runs. Each experiment is evaluated on a budget of $I=2000$ generative steps. In the main text, we display the total numbers after this budget of generative steps.

\subsection{Molecular Scaffolds}
\begin{figure*}
     \centering
     \begin{subfigure}[b]{0.30\textwidth}
         \centering
         \includegraphics[width=\textwidth]{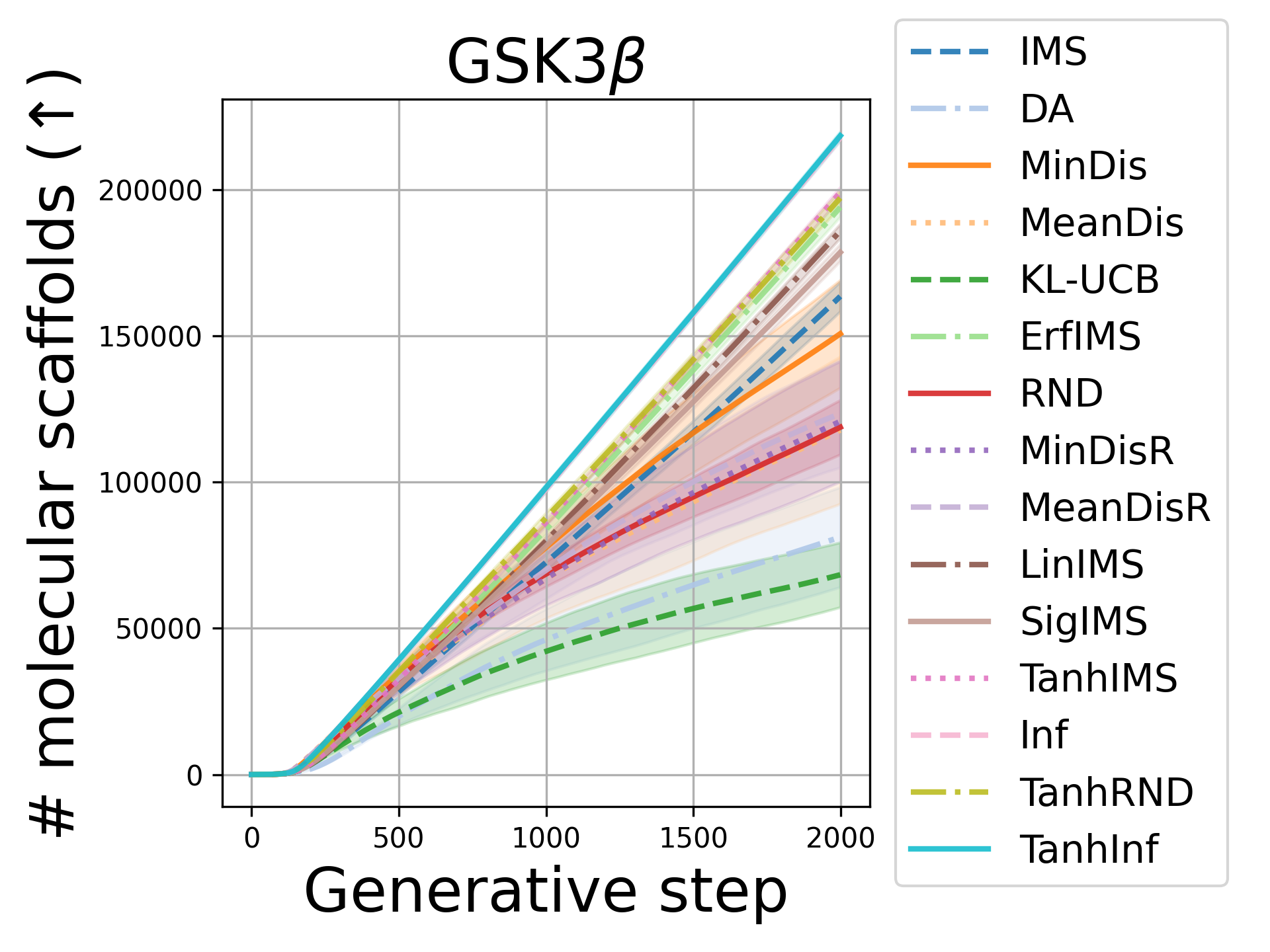}
         \caption{GSK3$\beta $}
         \label{fig:chem_per_step_gsk3b}
     \end{subfigure}
     \hfill
     \begin{subfigure}[b]{0.30\textwidth}
         \centering
         \includegraphics[width=\textwidth]{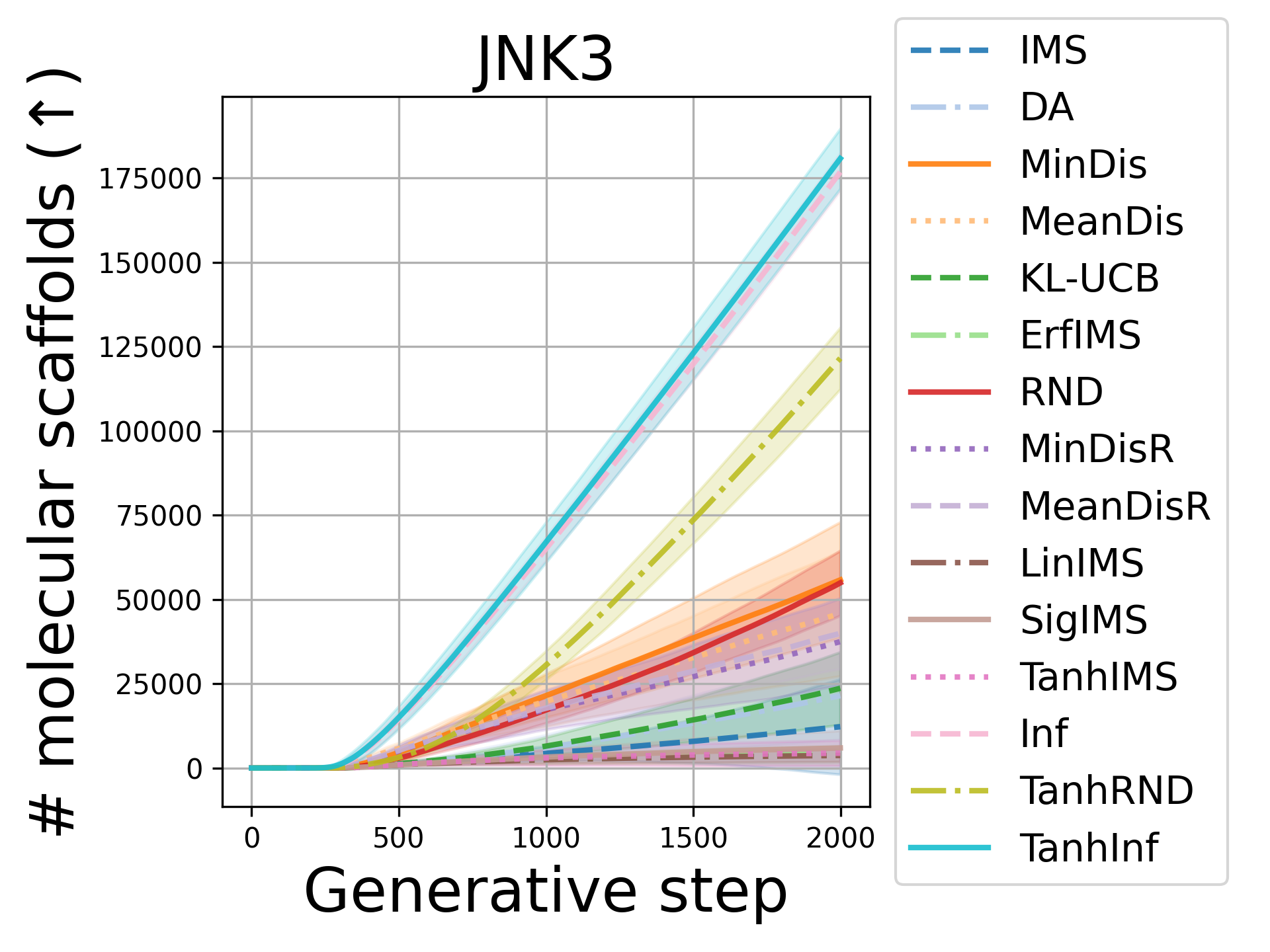}
         \caption{JNK3}
         \label{fig:chem_per_step_jnk3}
     \end{subfigure}
     \hfill
     \begin{subfigure}[b]{0.30\textwidth}
         \centering
         \includegraphics[width=\textwidth]{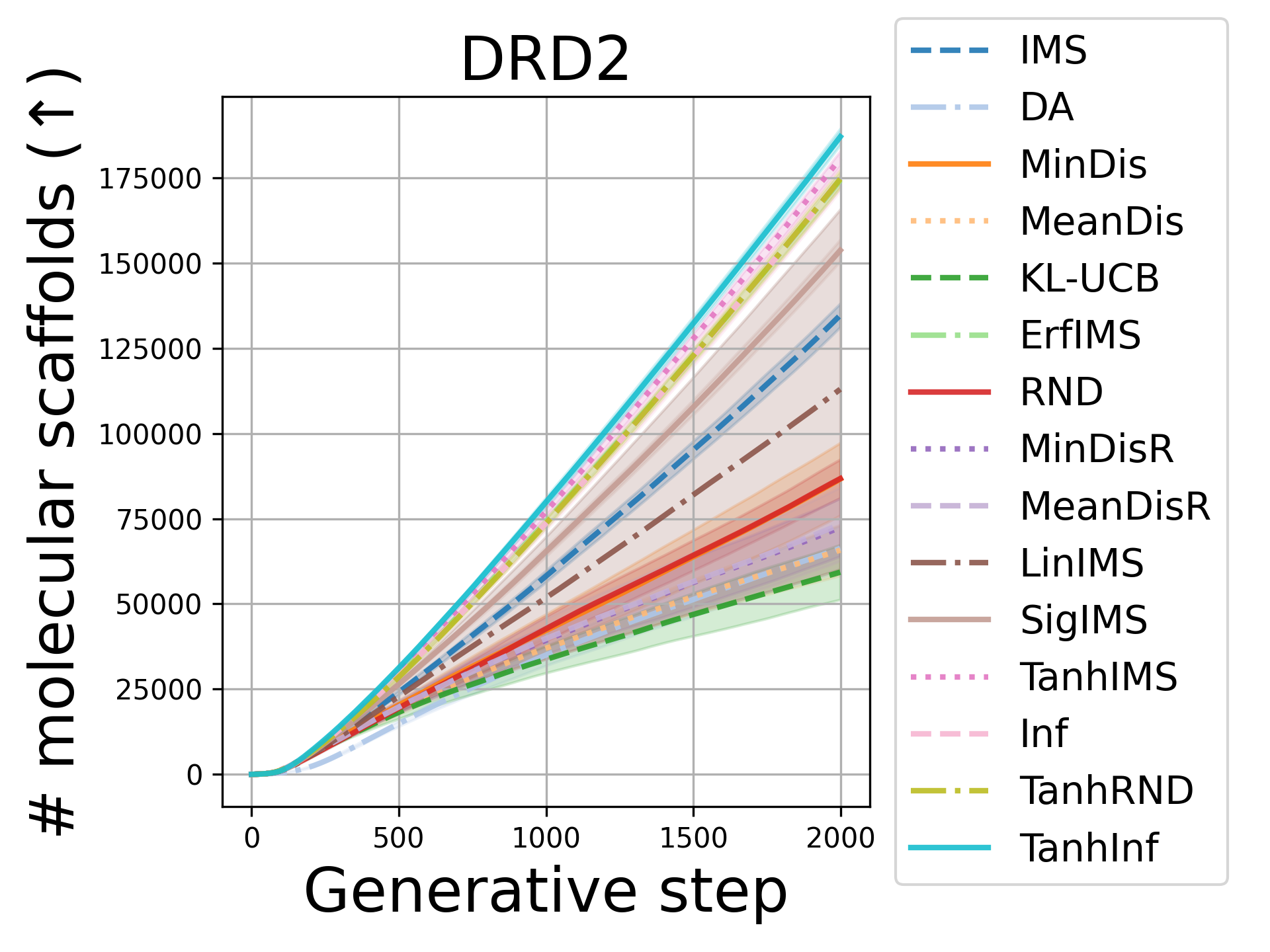}
         \caption{DRD2}
         \label{fig:chem_per_step_drd2}
     \end{subfigure}
        \caption{Total number of molecular scaffolds generated up to and including generative step $i$. Each line shows the mean over 20 reruns and the shaded region shows the sample standard deviation. }
        \label{fig:chem_per_step}
\end{figure*}

\Cref{fig:chem_per_step} shows the cumulative number of molecular scaffolds, across 20 independent runs, per generative step $i$. TanhInf is consistently the top diversity-aware reward function across all extrinsic reward functions (oracles). After 500 steps on the GSK3$\beta$ and JNK3 oracle, both Inf and TanhInf can generate significantly more molecular scaffolds per generative step than the other diversity-aware reward functions. For the GSK3$\beta$ experiments, the mean lines Inf and TanhInf almost fully overlap in terms of molecular scaffolds and, therefore, it is difficult to notice the line representing Inf. Moreover, TanhRND seems to be the third-best option in terms of the number of molecular scaffolds generated. In fact, in the JNK3 and DRD2 experiments, it can generate more molecular scaffolds than the other diversity-aware functions, except Inf and TanhInf; while it is among the top-performing options on GSK3$\beta$.

\subsection{Topological Scaffolds}
\begin{figure*}
     \centering
     \begin{subfigure}[b]{0.30\textwidth}
         \centering
         \includegraphics[width=\textwidth]{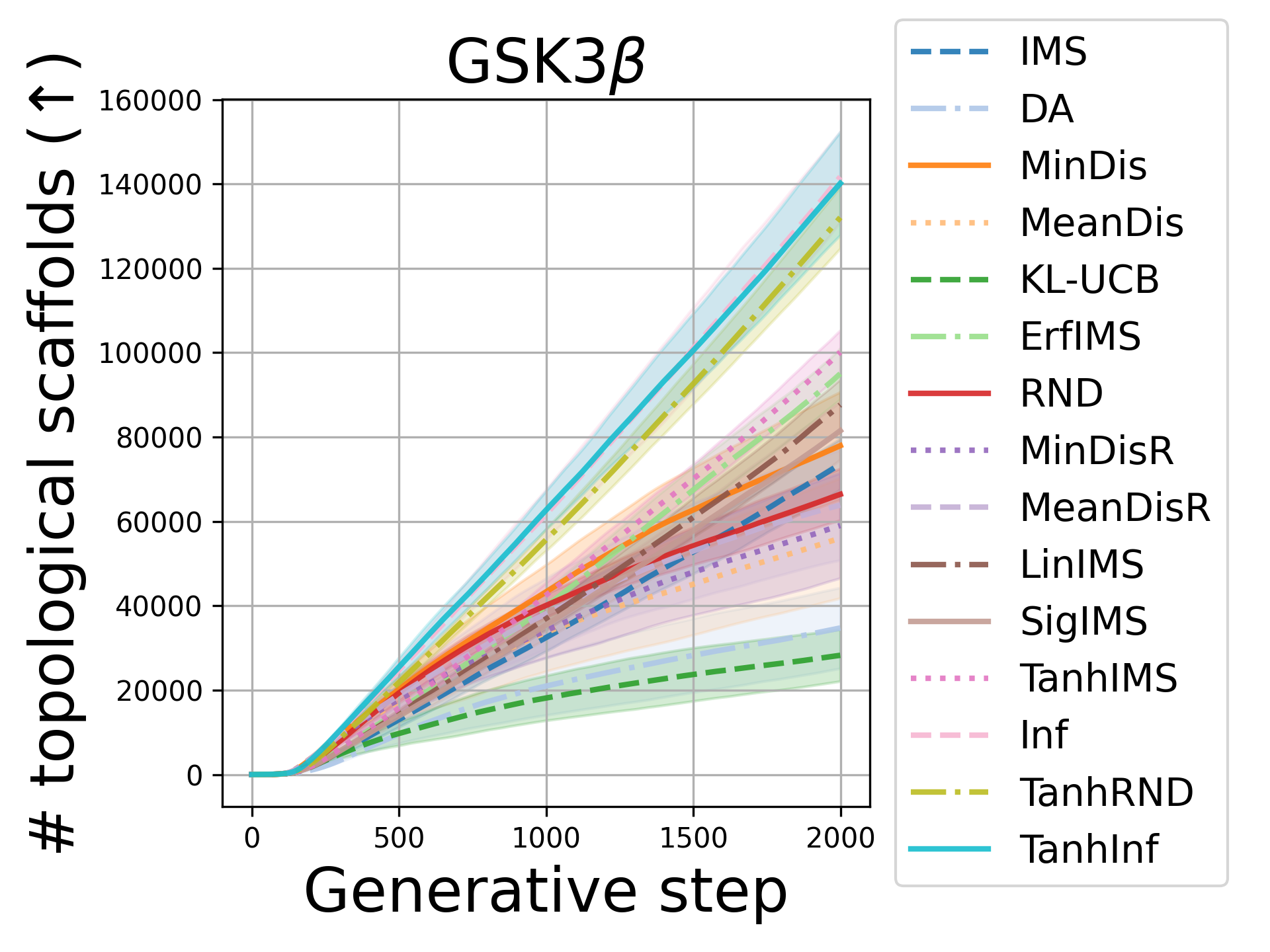}
         \caption{GSK3$\beta $}
         \label{fig:topo_per_step_gsk3b}
     \end{subfigure}
     \hfill
     \begin{subfigure}[b]{0.30\textwidth}
         \centering
         \includegraphics[width=\textwidth]{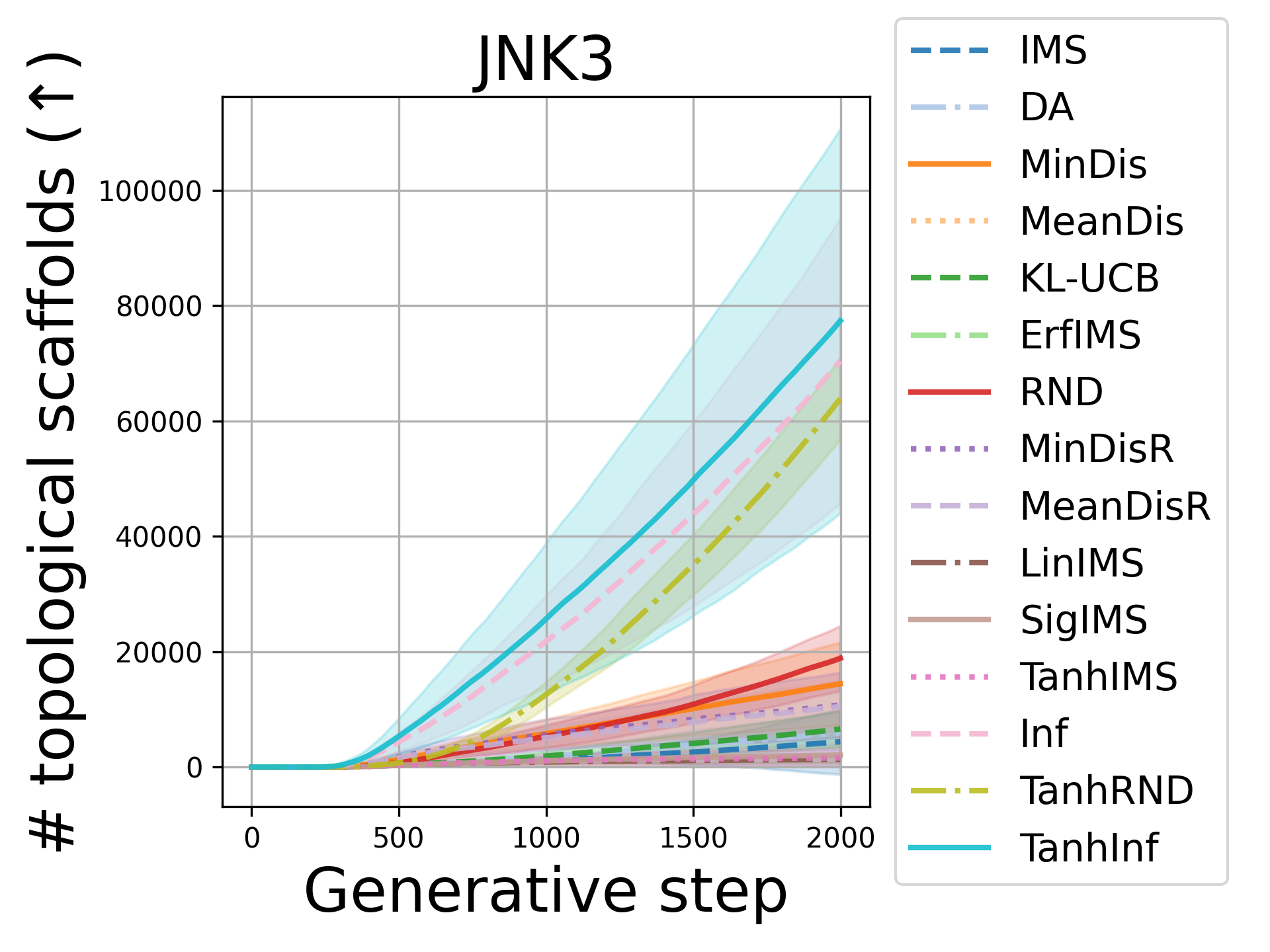}
         \caption{JNK3}
         \label{fig:topo_per_step_jnk3}
     \end{subfigure}
     \hfill
     \begin{subfigure}[b]{0.30\textwidth}
         \centering
         \includegraphics[width=\textwidth]{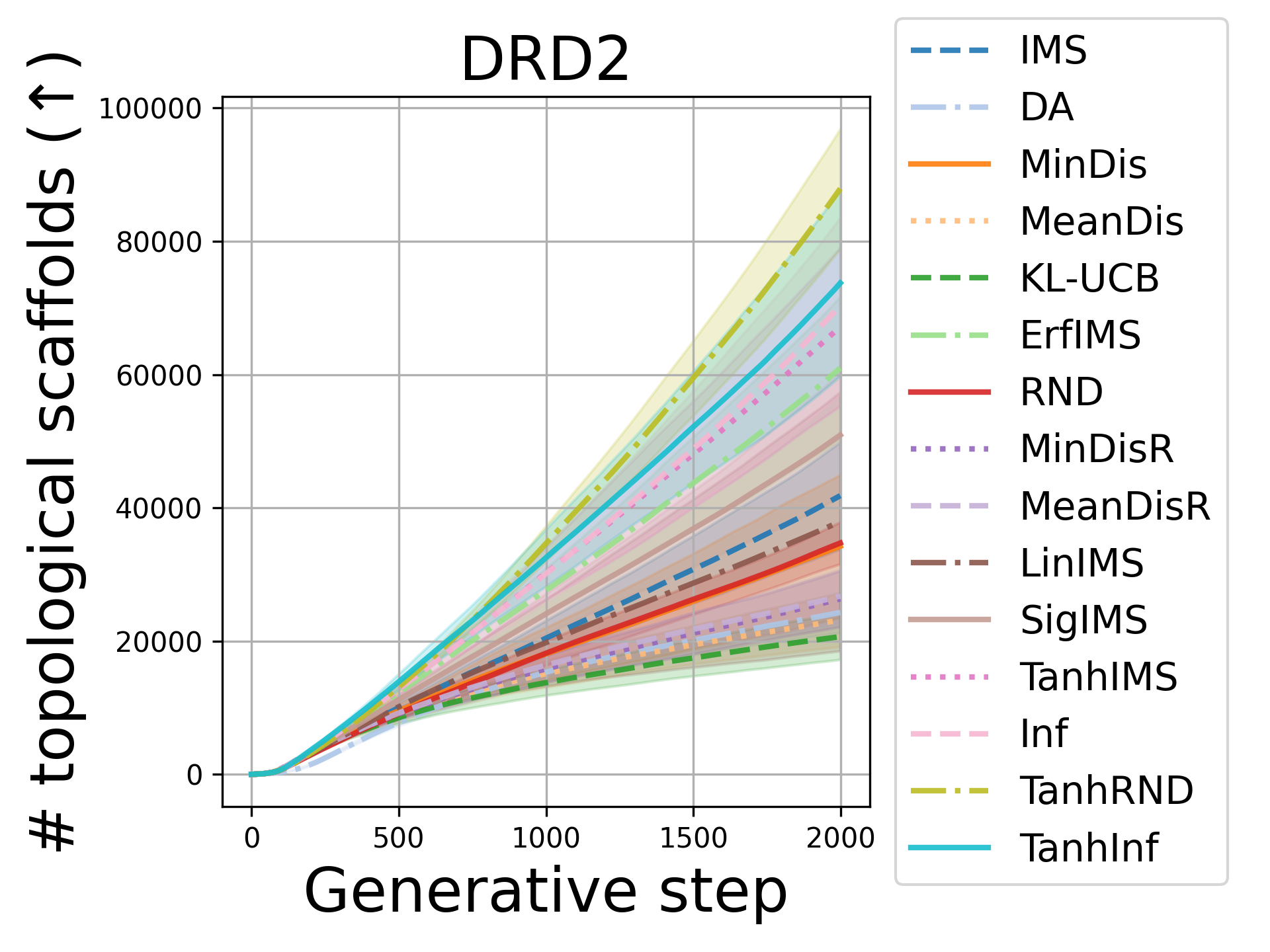}
         \caption{DRD2}
         \label{fig:topo_per_step_drd2}
     \end{subfigure}
        \caption{Total number of Topological scaffolds generated up to and including generative step $i$. Each line shows the mean over 20 reruns and the shaded region shows the sample standard deviation. }
        \label{fig:topo_per_step}
\end{figure*}

\Cref{fig:topo_per_step} shows the cumulative number of Topological scaffolds, across 20 independent runs, per generative step $i$. After around 750 generative steps, the diversity-aware reward functions TanhInf, TahnRND and Inf consistently generate more Topological scaffolds than the other functions.

\subsection{Diverse Actives}
\begin{figure*}
     \centering
     \begin{subfigure}[b]{0.3\textwidth}
         \centering
         \includegraphics[width=\textwidth]{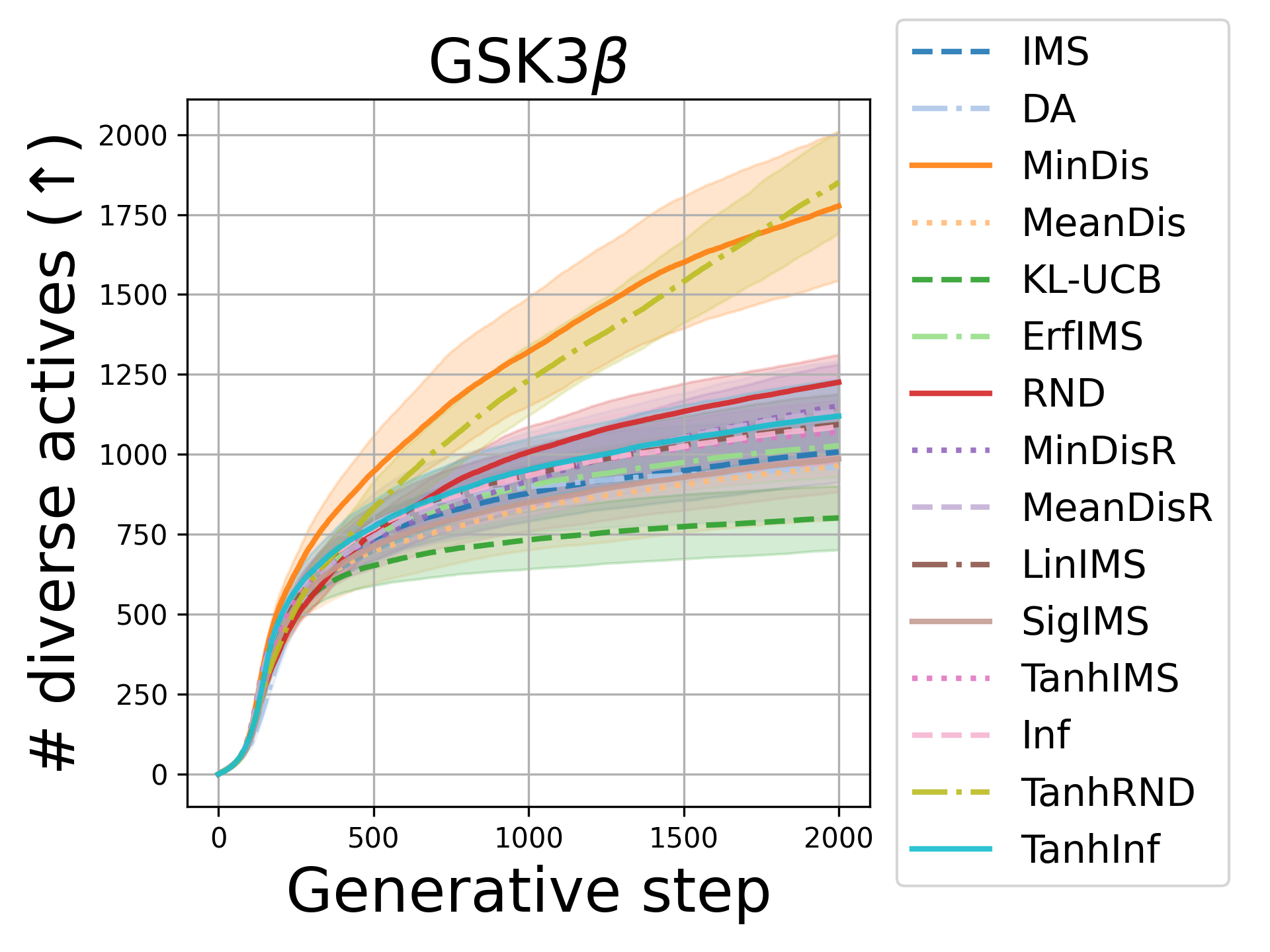}
         \caption{GSK3$\beta $}
         \label{fig:dh_per_step_gsk3b}
     \end{subfigure}
     \hfill
     \begin{subfigure}[b]{0.3\textwidth}
         \centering
         \includegraphics[width=\textwidth]{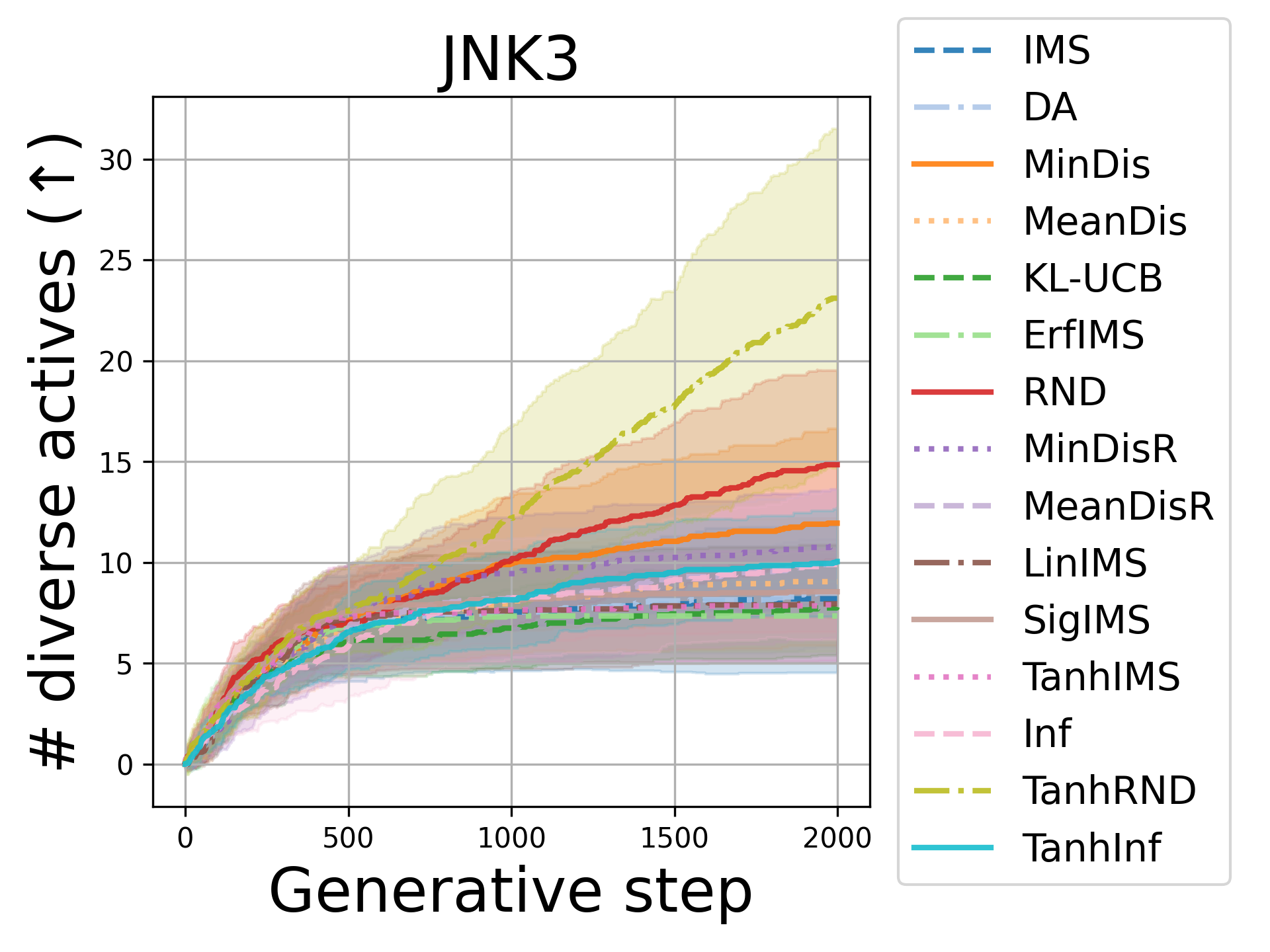}
         \caption{JNK3}
         \label{fig:dh_per_step_jnk3}
     \end{subfigure}
     \hfill
     \begin{subfigure}[b]{0.3\textwidth}
         \centering
         \includegraphics[width=\textwidth]{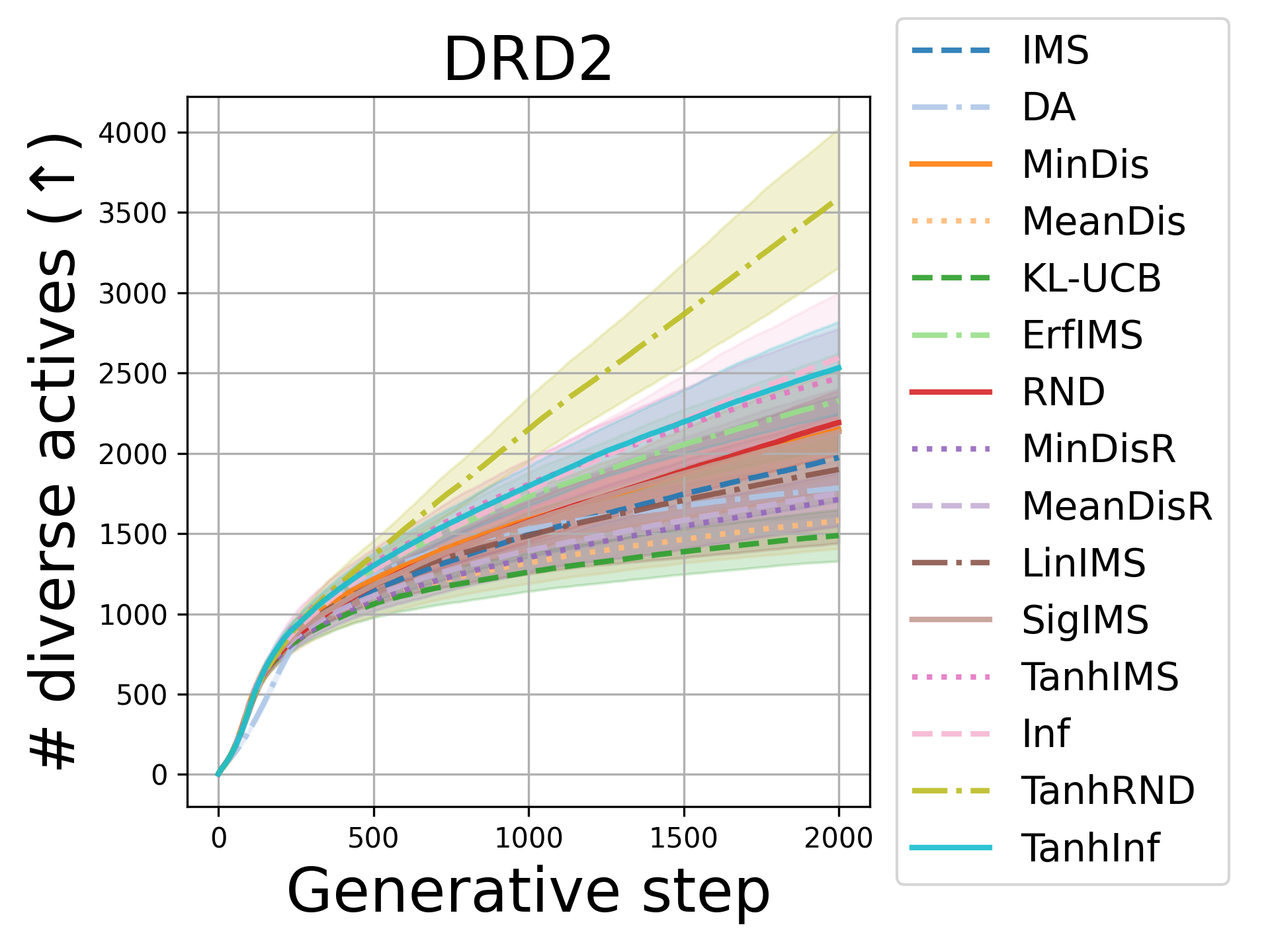}
         \caption{DRD2}
         \label{fig:dh_per_step_drd2}
     \end{subfigure}
        \caption{Total number of diverse actives generated up to and including generative step $i$. Each line shows the mean over 20 reruns and the shaded region shows the sample standard deviation. }
        \label{fig:dh_per_step}
\end{figure*}

\Cref{fig:dh_per_step} shows the cumulative number of diverse actives, across 20 independent runs, per generative step $i$. The diversity-aware reward function TanhRND can generate substantially more diverse activities per generative step $i$ across all oracles. 

\end{document}